\documentclass[journal]{IEEEtran}

\usepackage{amsmath,amsfonts}
\usepackage{algorithmicx}
\usepackage{algorithm}
\usepackage{array}
\usepackage[caption=false,font=normalsize,labelfont=sf,textfont=sf]{subfig}
\usepackage{textcomp}
\usepackage{stfloats}
\usepackage{url}
\usepackage{verbatim}
\usepackage{graphicx}
\usepackage{cite}
\usepackage{amsmath,amssymb,amsfonts}
\usepackage{graphicx}
\usepackage{textcomp}
\usepackage{xcolor}
\usepackage{hyperref}
\usepackage{svg}
\usepackage{multirow}
\usepackage{subcaption}
\usepackage{algorithm}
\usepackage{algpseudocode}
\usepackage{CJKutf8}
\usepackage{array}
\usepackage{amsmath,amssymb,amsfonts}
\usepackage{enumerate}
\usepackage{amsmath,amsfonts}
\usepackage{algorithmicx}
\usepackage{algorithm}
\usepackage{array}
\usepackage[caption=false,font=normalsize,labelfont=sf,textfont=sf]{subfig}
\usepackage{textcomp}
\usepackage{stfloats}
\usepackage{url}
\usepackage{verbatim}
\usepackage{graphicx}
\usepackage{cite}
\usepackage{amsmath,amssymb,amsfonts}
\usepackage{graphicx}
\usepackage{makecell}
\usepackage{textcomp}
\usepackage{xcolor}
\usepackage{hyperref}
\usepackage{svg}
\usepackage{multirow}
\usepackage{subcaption}
\usepackage{algorithm}
\usepackage{algpseudocode}
\usepackage{CJKutf8}
\usepackage{array}
\usepackage{fancyhdr} 
\usepackage{amsmath,amssymb,amsfonts}
\graphicspath{ {image/} }  
\def\BibTeX{{\rm B\kern-.05em{\sc i\kern-.025em b}\kern-.08em
    T\kern-.1667em\lower.7ex\hbox{E}\kern-.125emX}}

\makeatletter
\newcommand{\linebreakand}{%
  \end{@IEEEauthorhalign}
  \hfill\mbox{}\par
  \mbox{}\hfill\begin{@IEEEauthorhalign}
}
\makeatother
\IEEEoverridecommandlockouts                              

\begin{document}

\title{RobKiNet: Robotic Kinematics Informed Neural Network for Optimal Robot Configuration Prediction}

\author{Yanlong Peng, Zhigang Wang, Yisheng Zhang, Pengxu Chang, Ziwen He, Kai Gu, Hongshen Zhang, Ming Chen*
\thanks{Yanlong Peng, Yisheng Zhang, Pengxu Chang, Ziwen He and Kai Gu are with the School of Mechanical Engineering, Shanghai Jiao Tong University, Shanghai, China. (e-mail: \{me-pengyanlong, zys\_edward, 19119175490, heziwen, gukai0707\}@sjtu.edu.cn)}%
\thanks{Zhigang Wang is with the Intel Labs China, Beijing, China. (e-mail: zhi.gang.wang@intel.com )}%
\thanks{Hongshen Zhang is with the Faculty of Mechanical and Electrical Engineering, Kunming University of Science and Technology, Yunnan, China. (e-mail: hongshen@kust.edu.cn)}%
\thanks{Ming Chen, corresponding author, is with the School of Mechanical Engineering, Shanghai Jiao Tong University, Shanghai, China. (Address: No. 800 Dongchuan Road, Minhang District, Shanghai, China. e-mail: mingchen@sjtu.edu.cn )}%
}



\maketitle

\begin{abstract}


Task and Motion Planning (TAMP) is essential for robots to interact with the world and accomplish complex tasks. 
The TAMP problem involves a critical gap: exploring the robot's configuration parameters (such as chassis position and robotic arm joint angles) within continuous space to ensure that task-level global constraints are met while also enhancing the efficiency of subsequent motion planning. Existing methods still have significant room for improvement in terms of efficiency. Recognizing that robot kinematics is a key factor in motion planning, we propose a framework called the Robotic Kinematics Informed Neural Network (RobKiNet) as a bridge between task and motion layers. RobKiNet integrates kinematic knowledge into neural networks to train models capable of efficient configuration prediction. We designed a Chassis Motion Predictor(CMP) and a Full Motion Predictor(FMP) using RobKiNet, which employed two entirely different sets of forward and inverse kinematics constraints to achieve loosely coupled control and whole-body control, respectively. Experiments demonstrate that CMP and FMP can predict configuration parameters with 96.67\% and 98\% accuracy, respectively. That means that the corresponding motion planning can achieve a speedup of 24.24x and 153x compared to random sampling. Furthermore, RobKiNet demonstrates remarkable data efficiency. CMP only requires 1/71 and FMP only requires 1/15052 of the training data for the same prediction accuracy compared to other deep learning methods. These results demonstrate the great potential of RoboKiNet in robot applications.

\end{abstract}

\begin{IEEEkeywords}
AI-Based Methods, Kinematics, Differentiable Programming.
\end{IEEEkeywords}

\section{Introduction}

\begin{figure*}[h]
    \centering
    \includegraphics[scale=0.18]{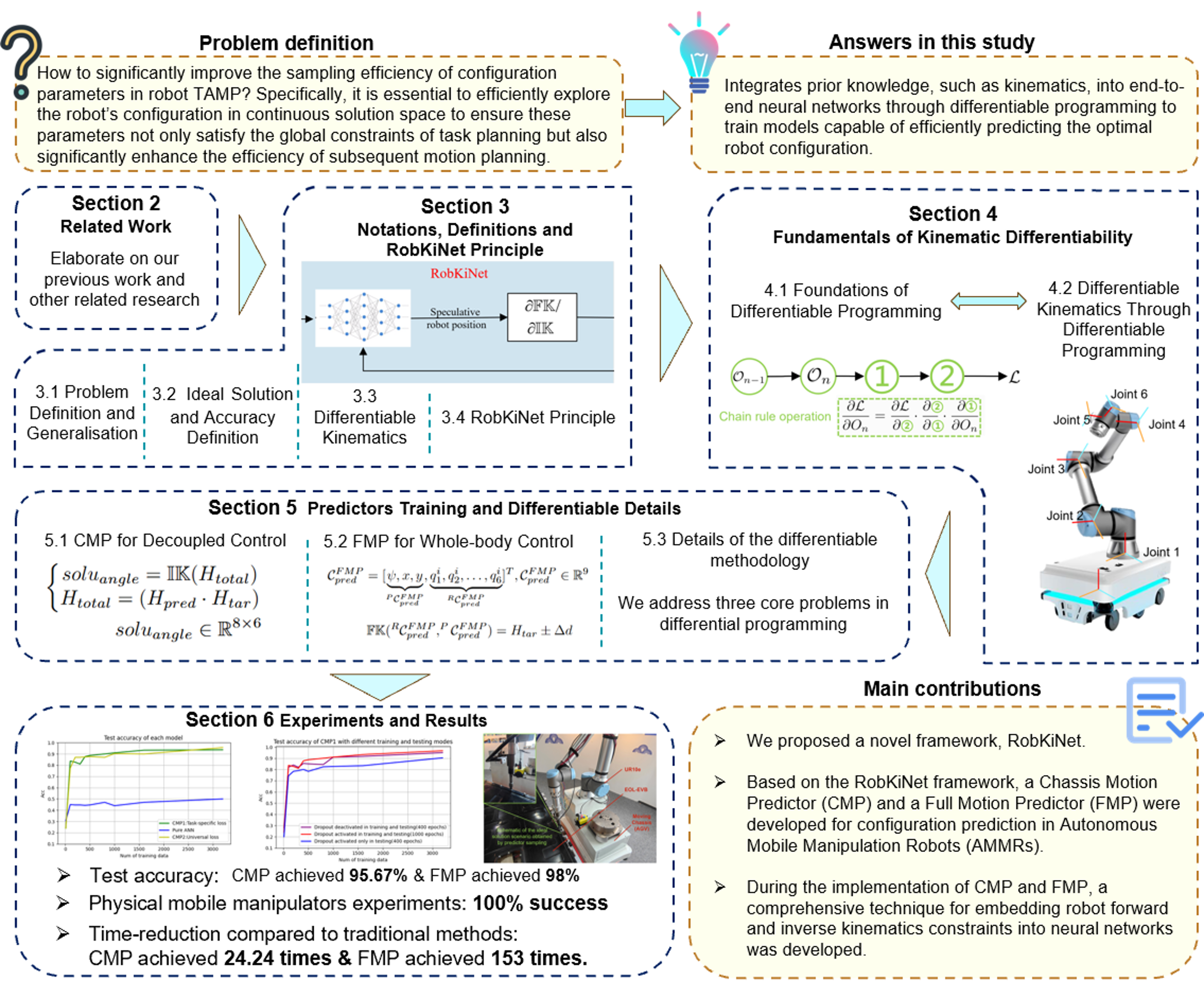}
    \caption{Overview of this paper's issues, ideas, chapter relationships, and main contributions.}
    \vspace{-2mm}
    \label{FIG:Overview}
\end{figure*}

Robot manipulation in real-world environments requires seamless integration of high-level task reasoning and low-level motion execution\cite{garrett2021integrated, munawar2018maestrob}. Task and Motion Planning (TAMP) frameworks have emerged as a powerful paradigm for addressing this challenge \cite{garrett2021integrated, primitive, garrett2018ffrob}, where task planners determine action sequences to achieve goals, while motion planners generate the corresponding trajectories. 

Specifically, the search at the task layer focuses on the abstract subsequences that should be executed to achieve the target state, such as subtask planning in long-horizon tasks \cite{sequence} or complex dynamic manipulation tasks \cite{ 88, primitive}. Speed improvements and near-optimal heuristic searches can be achieved by designing corresponding guided planners that can be learned  \cite{9} or efficient neural planners \cite{7,10}. 
On the other hand, the search at the motion layer is more focused on determining the specific parameter configurations of the entire robotic system when executing a particular action. The selected parameter configurations need to not only satisfy the current task's kinematic requirement but also ensure a feasible solution for the future task based on these parameter configurations \cite{matsuo2022deep,hanheide2017robot}.

However, the interface between these two planning layers remains a significant bottleneck in practical applications \cite{8, 88}. A key challenge in TAMP systems is determining viable robot configuration that satisfy both task-level objectives and motion-level constraints, especially for systems with redundant degrees of freedom, which is common in real-world scenarios\cite{zhang2023end}\cite{bouhsain2023learning}\cite{Navigation}. It involves exploring and selecting from a potentially infinite solution space to ensure the correctness of the execution of related tasks. 

In order to find viable robot configuration that are suitable for both the task level and the motion level, traditional approaches often rely on adding constraints or random sampling to find feasible configurations. A classic example is the inverse kinematics solution for a 7-degrees of freedom (DOF) robotic arm \cite{dou2022inverse, tian2021analytical}, where joint characteristics are constrained according to specific tasks to reduce the motion dimensions and narrow the continuous solution space. However, constraints are often difficult to determine when dealing with the aforementioned complex problems and inefficient. Random sampling \cite{sucan2012open,pddlstream} generate possible parameter configurations and check whether kinematic constraints satisfy all related tasks. However, this approach often becomes inefficient due to the large number of invalid constraint checks, despite many efforts to improve the success rate of sampling \cite{kingston2018sampling}.
When a motion planner fails to find a trajectory that satisfies the task planner's requirements, the system must backtrack and explore alternative solutions, significantly increasing planning time and reducing overall system reliability. 

\textbf{Therefore, the challenge addressed in this study is how to achieve efficient end-to-end prediction of the robot's configuration parameters while strictly adhering to the constraints of task planning and significantly enhancing the efficiency of subsequent motion planning. This challenge becomes especially urgent when the robot has redundant degrees of freedom and an infinite number of potential solutions exist within a continuous solution space.}

To address this critical challenge, recognizing the pivotal role of robot kinematics in TAMP,  we have proposed a framework named Robotic Kinematics Informed Neural Network (RobKiNet), a neural network-based framework that serves as an intelligent bridge between task and motion planning. 
Our approach learns to predict kinematically feasible robot configuration directly from task goals, effectively reducing the gap between abstract task specifications and concrete motion constraints. By integrates kinematic constraints into the training of end-to-end networks, RoboKiNet internalized and learned physical kinematic limitations. This significantly reduces the need for expensive iterative planning while maintaining high success rates in real-world manipulation tasks.

\begin{figure*}[h]
    \centering
    \includegraphics[scale=0.16]{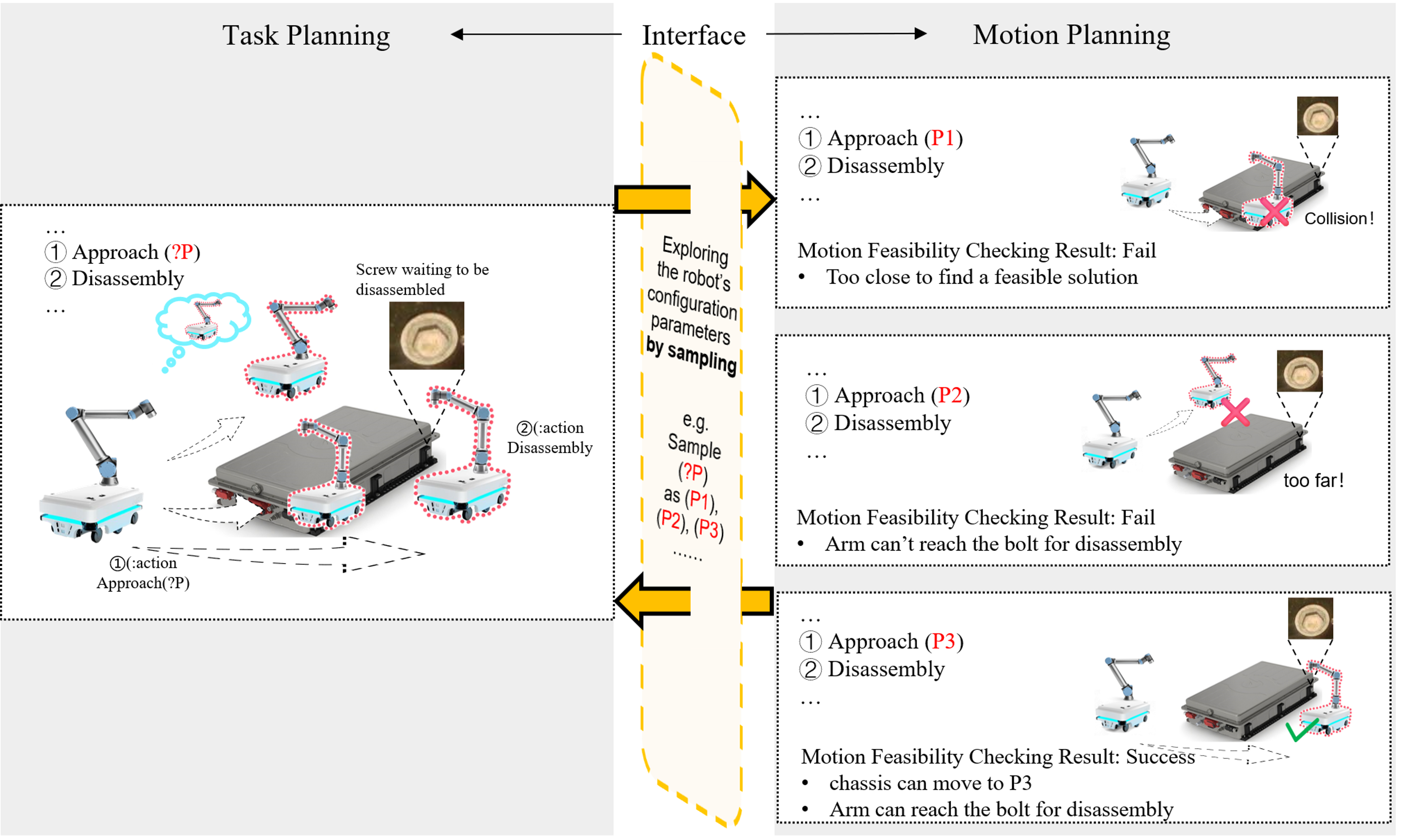}
    \caption{Example—Simplified TAMP framework for disassembling a screw on an EOL-EVB. (Left) Task Planning Level: The task planner, based on operational requirements (e.g., disassembling a particular bolt), outputs a sequence of action primitives (such as Approach, Disassembly, etc.) that guide the robot from its current state to the goal state. (Middle) Interface Between Task Planning and Motion Planning: This essential interface uses sampling methods to search for the robot’s configuration parameters and outputs them to the motion planning level. (Right) Motion Planning Level: The robot’s configuration parameters predicted by the interface—such as chassis position and joint angles of the robotic arm—are used to ensure the feasibility of execution. In this study, the predicted parameters from the interface meet both the global constraints of task planning (e.g., ensuring the robot is positioned near the target screw during the Approach primitive) and significantly enhance the efficiency of subsequent motion planning (ensuring feasible kinematic solutions).}
    \vspace{-2mm}
    \label{FIG:1}
\end{figure*}

In this study, we explore RobKiNet in the field of mobile robots \cite{1,6}, which requires deciding high-quality robot configuration parameters based on the constraints of task planning in an infinite continuous solution space for subsequent motion planning. The paper is organized as follows(Figure \ref{FIG:Overview}). After outlining the related work (Section \ref{Related Work}), we formally give the notations, define the problem, and illustrate the RobKiNet principle in Section \ref{Notation and Definitions}. Then, we provide some fundamental knowledge and explain differentiable kinematics in Section \ref{fundamental knowledge}. 
Next, we elaborate on our two attempts based on the RobKiNet. 
We use inverse kinematics (IK) as a physical constraint criterion for the control decoupled from the mobile robot in the chassis motion predictor (CMP) and forward kinematics (FK) as a physical constraint criterion for the control of the whole body \cite{minniti2019whole, li2020design, zhang2023end} in the full motion predictor (FMP). 
After that, we detail the core methods of using differentiable physical constraints to train end-to-end networks, which apply to all non-differentiable robotics constraints (Section \ref{RPSN Training Based on Differentiable Arithmetic Process Guidance}). Finally, we present the experimental details and real-world results (Section \ref{Experiments and Results}).

The main contributions of this paper are as follows:
\begin{enumerate}
\item We propose a novel framework, RobKiNet, which leverages differentiable programming to deeply integrate robot kinematics as physical constraints within neural networks, effectively constraining the output solution space and gradients to train an efficient and robust robot configuration prediction model. This approach enables the neural network to replace traditional sampling methods, achieving rapid configuration selection while simultaneously satisfying complex constraints from both task planning and motion planning layers, paving the way for new approaches in robot learning and operation.
\item Based on the RobKiNet framework, we developed a Chassis Motion Predictor (CMP) and a Full Motion Predictor (FMP) for autonomous mobile manipulation robots (AMMRs) equipped with a chassis and a six-degree-of-freedom robotic arm. These predictors output valid configurations directly in an end-to-end manner, enabling configuration prediction for both the chassis and the entire AMMR. Both predictors are available for download, facilitating practical applications.
\item During the implementation of CMP and FMP, a comprehensive technique for embedding robot forward and inverse kinematics constraints into neural networks was developed. This demonstrated the significant potential of differentiable programming techniques in the field of robot TAMP, providing valuable insights and references for the development and optimization of other differentiable computing frameworks.
\end{enumerate}

\section{Related Work}\label{Related Work}

\subsection{Our Previous Work}

In our previous research, we built a NeuroSymbolic \cite{NS-start1, NS-start2} Task and Motion Planning System(NeuroSymbolic TAMP\cite{zhanghengwei2,peng2024revolutionizing, zhanghengwei,zhangyisheng}) to control an Autonomous Mobile Manipulation Robot(AMMR) to accomplished an unstructured industrial task, continuous disassembly of end-of-life electric vehicle batteries (EOL-EVBs) screws. 

In such TAMP systems, task planning involves searching for the optimal sequence of subtasks in a high-level symbolic space to transition from the current state to the goal state. These subtasks typically require the robot to perform a series of motion primitives that each necessitate individual motion planning. However, there exists a essential interface between symbolic task planning and motion planning in Cartesian or configuration space(Figure \ref{FIG:1} (Middle)). This interface requires efficient sampling of robot configuration parameters to ensure compliance with task-level global constraints, while also considering constraints imposed by motion planning.

For instance, positioning the chassis too close to the battery may lead to kinematic singularities or no kinematic solution for the robot, whereas placing it too far could put the object outside the robot arm’s reachable workspace (Figure \ref{FIG:1} (Right)). In this context, efficient sampling of the chassis position is critical, as it takes place after the “Approach” command at the task planning level and before path planning and robotic arm kinematic computations. Its selection takes into account the task constraints while preparing for motion planning.

In this study, we will use the RobKiNet to bridge this gap for optimal robot configuration prediction, which benefits not only our previous research but also other research especially for systems with redundant degrees of freedom.

\subsection{Exploring Robot Configuration Parameter}

Since the solution space of the redundant degrees of freedom problems is continuous and lacks an explicit equation, the most common methods to explore the robot's configuration parameters are predefined relative nominal parameters or perform sampling.

Contemporary pre-programmed methods often rely on a fixed nominal parameter linked to the target to define the configuration parameter, such as chassis position, a practice commonly observed in analogous tasks\cite{AMMR1, disassembly_cell}. However, pre-programming fails in a highly dynamic environment with extensive workspace, diverse task types, and irregular work scenarios. Further, most mobile manipulators today have a loosely coupled chassis and robotic arm. The independent chassis controllers make it difficult to perform whole-body control from the navigation level. For these reasons, we cannot rely on predefined approaches to solve the parameter selection for the robotic arm and the chassis, as tasks are not fixed.

In the motion planning task about the chassis, we need to know both the current chassis position and the intended target position for chassis motion in advance to complete the path planning and robot kinematics simulation. 
It's common to employ random sampling(RS) \cite{pddlstream} or efficient biased sampling(EBS)\cite{1232271}, but this can lead to uncertainty about whether the sampled position is truly effective for task completion. Hence, conducting forward and inverse kinematics simulations of the robot in the simulator becomes necessary, which consumes significant computational resources.
The inefficiency of randomly sampling and simulating, together with the gap between simulation results and the real situation, are the key bottlenecks preventing robots from moving towards agility.

Furthermore, when considering the configuration parameters of the robotic arm, additional kinematic constraints must be accounted for \cite{bouhsain2023learning}. A common approach is to extend the DH matrix to include the chassis as a mobile joint within the overall configuration \cite{zhang2023end}, followed by solving through kinematic analysis. Another method involves using the Jacobian matrix for iterative solving in specific robotic arm configurations \cite{chen2017tracking}. However, these approaches often face challenges in solving complex analytical solutions or can only identify a single feasible solution within the solution space, limiting their ability to accurately and efficiently explore the potential solution space. The exploration of robot configuration parameters should incorporate more efficient methods, such as leveraging human prior knowledge.

\subsection{Deep learning for TAMP}

Considering the substantial benefits neural networks have demonstrated in learning from diverse data types, many researchers are trying to apply deep learning to TAMP. Regression in deep learning refers to a type of supervised learning task where the goal is to predict a continuous output variable based on input features. It is a natural candidate for TAMP. The advantage of it is that it can constrain and guide TAMP using known datasets \cite{7, 10, matsuo2022deep}. However, this method faces the challenge of obtaining large amounts of real-world physical interaction data. In addition, the inherent characteristic of motion planning being a one-to-many mapping further increases the difficulty of convergence for regression methods. 

Deep reinforcement learning (DRL) \cite{haarnoja2024learning, 9699963, zhang2021hierarchical} is another widely used deep learning method for TAMP. It focuses on training agents to make a sequence of decisions by interacting with an environment to maximize cumulative rewards. While DRL has shown remarkable success in various applications\cite{zhang2019efficient,luperto2020robot}, it also comes with efficiency challenges. DRL algorithms often require a large number of interactions with the environment to learn effective policies. This can be particularly problematic in real-world applications where data collection is expensive or time-consuming.

Physics-informed neural networks (PINNs) can effectively incorporate physical equations as constraints into the neural network training process \cite{chiu2022can, 9403414}. This provides a new approach to solving TAMP planning problems using deep learning. However, PINNs cannot be directly applied to the domain of robot kinematics because they require prior knowledge in the form of Partial Differential Equations(PDEs).

The approach taken by PINNs is inspiring, and we aim to take it a step further in the field of robotics. By incorporating complex constraints as prior knowledge beyond just partial differential equations, we can enable efficient sampling within a narrowed solution space. 

We aim to develop a framework that incorporates robot kinematic knowledge as constraints within a neural network to predict robot configuration parameters. This framework aims to provide an efficient solution for motion planning—specifically, the sampling of infinite solutions in continuous space—while satisfying the global constraints required by task planning.

\begin{table*}[ht]
\centering
\caption{Notations and Definitions} \label{label: Notations and definitions}
\renewcommand{\arraystretch}{1.5} 
\begin{tabular}{|c|c|}
\hline
\textbf{Notations} & \textbf{Definitions} \\
\hline
$\mathcal{C}$ & Configuration space, representing all possible configurations of the robot system.  \\
\hline
$\mathcal{C}_{base}$/ $\mathcal{C}_{arm}$ & Configuration space of the mobile chassis/manipulator's joints.\\
\hline
$\mathcal{X}$ & Task-specific solution space, a subset of $\mathcal{C}$ that satisfies a specific task’s constraints. \\
\hline
$\mathcal{C}^*$ & Globally constrained solution space, a subset of $\mathcal{X}$ that satisfies additional global constraints. \\
\hline
$\mathcal{C}^*_{base}$ & Optimal solutions/Ideal solution of the chassis.\\
\hline
$\boldsymbol{\theta}$ & Manipulator joints' angle matrix.\\
\hline
$\boldsymbol{\theta}^*$ & Optimal solutions/Ideal solution of the manipulator joints' angle.\\
\hline
$\mathcal{C}_{pred}$ &  Estimated configuration through forward propagation of the end-to-end network.\\
\hline
$\mathcal{C} _{pred}^{CMP}$/ $\mathcal{C} _{pred}^{FMP}$  & Output of the CMP or the FMP.\\
\hline
$^{P}\mathcal{C}_{pred}^{FMP}$/ $^{R}\mathcal{C}_{pred}^{FMP}$ & Sample for the chassis/Sample for the manipulator in FMP. Decoupled and independent in computation.\\
\hline
$\mathcal{T}_{target}$ & The position and orientation of the target object.\\
\hline
$\mathcal{L}$ & The real-valued space representing loss metrics, guiding the training process of the end-to-end network.\\
\hline
$\mathcal{W},\mathcal{W}_{1}$ & Manipulator’s workspace(Cartesian space) at its current different $\mathcal{C}_{pred}$.\\
\hline
$(w, b)$ & Refers to the weights and biases in the neural network. \\
\hline
$\mathbb{IK}/ \mathbb{FK}$ & Differentiable kinematics computation
engines obtained through differentiable programming.\\
\hline
$\mathcal{O}$, $\mathcal{O}_{1}, \dots, \mathcal{O}_{n}$ & Different operators in the computation graph that represent differentiable operations.\\
\hline
$H_{pred}, H_{tar}$ & Homogeneous transformation matrix form of $\mathcal{C}_{pred}$, $\mathcal{T}_{target}$\\
\hline
$H_{total}$ & Represents the coordinate relationship between the end joint and the base(homogeneous transformation matrix).\\
\hline
$\Delta d$ & The prediction error of the FMP.\\
\hline
${(:action} $ Approach & The PDDL\cite{1998pddl,pddlstream} form of the action primitive ``Approach" in our NeuroSymbolic TAMP Framework.\\
\hline
\end{tabular}
\label{tab:symbols}
\end{table*}

\section{Notations, Definitions and RobKiNet Principle}\label{Notation and Definitions}
This section will provide some commonly used symbols and formal definitions. We rigorously describe the concepts that appear frequently in this paper and define the problems and evaluation metrics of interest in this study. For ease of reading, we have provided a symbol table (Table \ref{label: Notations and definitions}) to clarify potentially confusing notations and definitions. Further, we explain the principles of the RobKiNet in general terms using a contrast with traditional methods.

\subsection{Problem Definition and Generalisation}

In robotic systems with $n$ degrees of freedom, the configuration space $\mathcal{C} \subseteq \mathbb{R}^n$ represents the set of all possible configurations the system can achieve. $\mathcal{C}$ is the most expansive space, encompassing every feasible configuration defined solely by the system's mechanical structure. However, when performing a specific task, the solution space is reduced to a subset $\mathcal{X} \subseteq \mathcal{C}$, constrained by the requirements of that particular task. The task-specific solution space 
$\mathcal{X}$ consists of an infinite number of valid configurations that all satisfy the given task's constraints, forming a continuous space.

Importantly, when additional global constraints are introduced, such as multi-tasks kinematic constraints or environmental restrictions, the solution space becomes further refined, reducing to $\mathcal{C}^* \in \mathcal{X}$. This represents the most restrictive subset, where each configuration satisfies not only the task-specific constraints but also the motion layers' constraints.
From a topological perspective, the space $\mathcal{C}^*$ should ideally form an open set within $\mathcal{X}$. This ensures robustness, as small perturbations around any point in $\mathcal{C}^*$ will remain within $\mathcal{X}$, meaning that slight deviations in configuration do not invalidate the solution. This hierarchical structure - from the overall configuration space $\mathcal{C}$ to the task-specific solution space 
$\mathcal{X}$ and finally to the globally constrained solution space $\mathcal{C}^*$ - allows for a rigorous and robust framework for understanding how robotic systems operate within complex environments.

The problem now shifts to sampling within the infinite, continuous space $\mathcal{X}$  to identify the optimal solution space $\mathcal{C}^* \in \mathcal{X}$, guided by a loss function $\mathcal{L}: \mathcal{C} \to \mathbb{R}$, where $\mathbb{R}$ denotes the real-valued space of loss metrics, distinct from the configuration space $\mathbb{R}^n$. Each configuration in $\mathcal{X}$ is evaluated by this loss function, representing how well the solution aligns with all global constraints. The challenge lies in efficiently exploring $\mathcal{X}$, selecting a set of candidate solutions, and identifying the optimal configuration in $\mathcal{C}^*$ that minimizes the loss $\mathcal{L}(\mathcal{C})$ while ensuring that all system constraints are satisfied.

In our example of the mobile robot (Figure \ref{FIG:1}), consider the configuration space $\mathcal{C}_{base}$ for the mobile chassis, typically represented as a continuous space(e.g., $\mathbb{R}^2$ or $\mathbb{R}^3$). Let $\mathcal{C}_{arm}$ denote the joint configuration space of the manipulator's joints, described by the vector of joint angles $\boldsymbol{\theta} \in \mathcal{C}_{arm}$. For a manipulator with 6-DOF, the $\boldsymbol{\theta}$ can be represented as: 
\begin{equation}
\begin{aligned}
    \boldsymbol{\theta} = \left(q^{i}_{j} \right)_{8 \times 6}, \quad i = 1, 2, \dots, 8 \text{ and }  j = 1, 2, \dots, 6 \label{eq:66} 
\end{aligned}
\end{equation}

This represents eight sets of inverse kinematics solutions, where each $q^{i}$ denotes the $i-th$ set of solutions, consisting of the values for six joints.
The task involves finding configurations $\mathcal{C}^* \in \mathcal{C}_{base}$ for the system such that the target object is within the manipulator's workspace $\mathcal{W}$ and the manipulator's inverse kinematics solution exists.

The problem can be formalized as follows:

\noindent \textbf{Chassis Position Validity Condition}: The system configurations in $\mathcal{C}^*$ must allow the end-effector of the manipulator to reach the target object position and orientation $\mathcal{T}_{target}$. This requires finding a chassis position $\mathcal{C}_{base}^*$ and a corresponding arm configuration $\boldsymbol{\theta}^*$ such that the forward kinematics function  $\mathcal{F}(\mathcal{T}_{base}^*, \boldsymbol{\theta}^*)$ maps to $\mathcal{T}_{target}$:

\begin{equation}
\mathcal{F}(\mathcal{T}_{base}^*, \boldsymbol{\theta}^*) = \mathcal{T}_{target}
\end{equation}

where $\mathcal{F}$ represents the forward kinematics mapping the chassis and arm configurations to the workspace.

\noindent \textbf{Existence of Inverse Kinematics Solution}: For a given chassis position $\mathcal{C}_{base}^*$, there must exist at least one arm configuration $\boldsymbol{\theta}^*$ such that the end-effector can achieve the desired $\mathcal{T}_{target}$. The set of all valid arm configurations for a given chassis position is defined as:
\begin{equation}
\mathcal{S} = \left\{ \boldsymbol{\theta}^* \in \mathcal{C}_{arm} \mid \mathcal{F}(\mathcal{C}_{base}^*, \boldsymbol{\theta}^*) = \mathcal{T}_{target} \right\}
\end{equation}

The goal is to find $\mathcal{C}_{base}^* \in \mathcal{C}_{base}$ such that $ S \neq \emptyset$.

Since $\mathcal{C}_{base}$ and $\mathcal{C}_{arm}$ are continuous spaces, the set of solutions $\mathcal{C}^*$ that satisfy the above conditions forms a continuous manifold, potentially with infinitely many solutions. This scenario illustrates the challenge of dealing with infinite solution sets in continuous spaces, where the exact enumeration of solutions is infeasible.

The motivation for this task is easily understood when we consider it from a human perspective. When a person needs to reach for a cup on a distant table, they instinctively move to an appropriate target position. Humans can achieve this effortlessly and without conscious thought because their brain inherently accounts for constraints such as the arm's workspace and singularities.

Furthermore, when performing TAMP on robots under prior constraints such as world models, the sampling problems in continuous solution spaces are abundant. Another common example is the grasping problem, where the grasp points are continuous for different objects, yet they are subject to directional constraints \cite{bouhsain2023learning}.

\subsection{Ideal Solution and Accuracy Definition}

We aim to solve this problem through end-to-end networks, and therefore, we need to define the ideal solution. The definition of the ideal solution does not rely on dataset mapping relationships but on known constraints, such as the CMP and the FMP defined under different kinematic constraints. The significance of defining the ideal solution lies in the need to guide and supervise backpropagation during network training. Our advantage lies in the fact that network training can be guided by known constraints, which saves a significant amount of computational resources.

Suppose $\mathcal{C}_{pred}$ is the estimated target position obtained by the robot through forward propagation of the end-to-end network.
Ideal solution space $\mathcal{C}^*$ is defined by satisfying the following conditions:

\noindent \textbf{Existence Condition}

The relationship between the target position and the workspace, as well as the decision on whether movement is required, is determined at the task planning level. However, it must always ensure that after motion planning:
\begin{equation}
\mathcal{T}_{target} \in \left\{\mathcal{W}_{1} \mid \mathcal{C}_{base}^* \in \mathcal{C}_{base} \right\}
\end{equation}

\noindent \textbf{Reachability Condition}

By using the CMP for decoupled control, the $\boldsymbol{\theta}^*$ is implicit.
By using the FMP for whole-body control, the $\boldsymbol{\theta}^*$ is an explicit result. Both of the situations satisfy the Chassis Position Validity Condition mentioned before and: 

\begin{equation}
\exists \boldsymbol{\theta}^* = q^{i}_{j}, q^{i}_{j} \in [-\pi, \pi]\label{eq:6} 
\end{equation}

Then the accuracy of the network in this research $ACC$ is defined as:
\begin{equation}
ACC = \frac{\sum_{i=1}^N \mathbb{I} (\mathcal{C}_{pred}, \mathcal{C}^*)}{N}
\end{equation}

where $\mathbb{I}$ is an indicator function. The function takes the value of 1 when the predicted value $\mathcal{C}_{pred}$ matches the ideal space $\mathcal{C}^*$, and 0 otherwise. $N$ is the total number of samples using random ways or the methodologies proposed in this research. 

This metric directly reflects a sampling method's ability to provide the ideal solution across N independent tasks, thereby indicating whether the end-to-end network guiding the robot has genuinely learned the physical constraints.
In the following experiments, we will compare random sampling/iteration  methods, traditional supervised/reinforcement learning methods, and the two predictors we proposed by the RobKiNet, with $ACC$ serving as the core evaluation metric.

\subsection{Differentiable Kinematics}
It is important to clarify that all mentions of ``Differentiable'' in this study refer specifically to ``Differentiable Kinematics through differentiable programming \cite{Differentiableprogramming, Differentiableprogramming2}.'' The differentiable kinematics here does not refer to the common practice of total differentiation for dynamics solving, but rather to differentiability at the computational graph level \cite{Differentiableprogramming, Differentiableprogramming3, Differentiableprogramming4}. This is a key method in our proposed RobKiNet for incorporating human prior knowledge and physical constraints into end-to-end networks for robots. Consequently, backpropagation is defined as follows:

\noindent \textbf{Differentiable Kinematics}:
differentiable programming $\{IK, FK\} \rightarrow \{ \mathbb{IK}, \mathbb{FK} \}$

\begin{equation}\label{Differentiable Kinematics}
\frac{\partial \mathcal{L}}{\partial (w, b)} = \frac{\partial \mathcal{L}}{\partial \mathbb{IK}/ \mathbb{FK}} \cdot \frac{\partial \mathbb{IK}/ \mathbb{FK}}{\partial y} \cdot \frac{\partial y}{\partial (w, b)}
\end{equation}

Here, we use $\{IK\}$, $\{FK\}$ to denote the processes of inverse kinematics and forward kinematics, respectively. $\mathbb{IK}$ and $\mathbb{FK}$ represent the differentiable kinematics computation engines obtained through differentiable programming methods discussed in this paper. $(w, b)$ refers to the weights and biases in the neural network. $y$ denotes the network prediction results.

\begin{figure*}[t]
    \centering
    \includegraphics[scale=0.13]{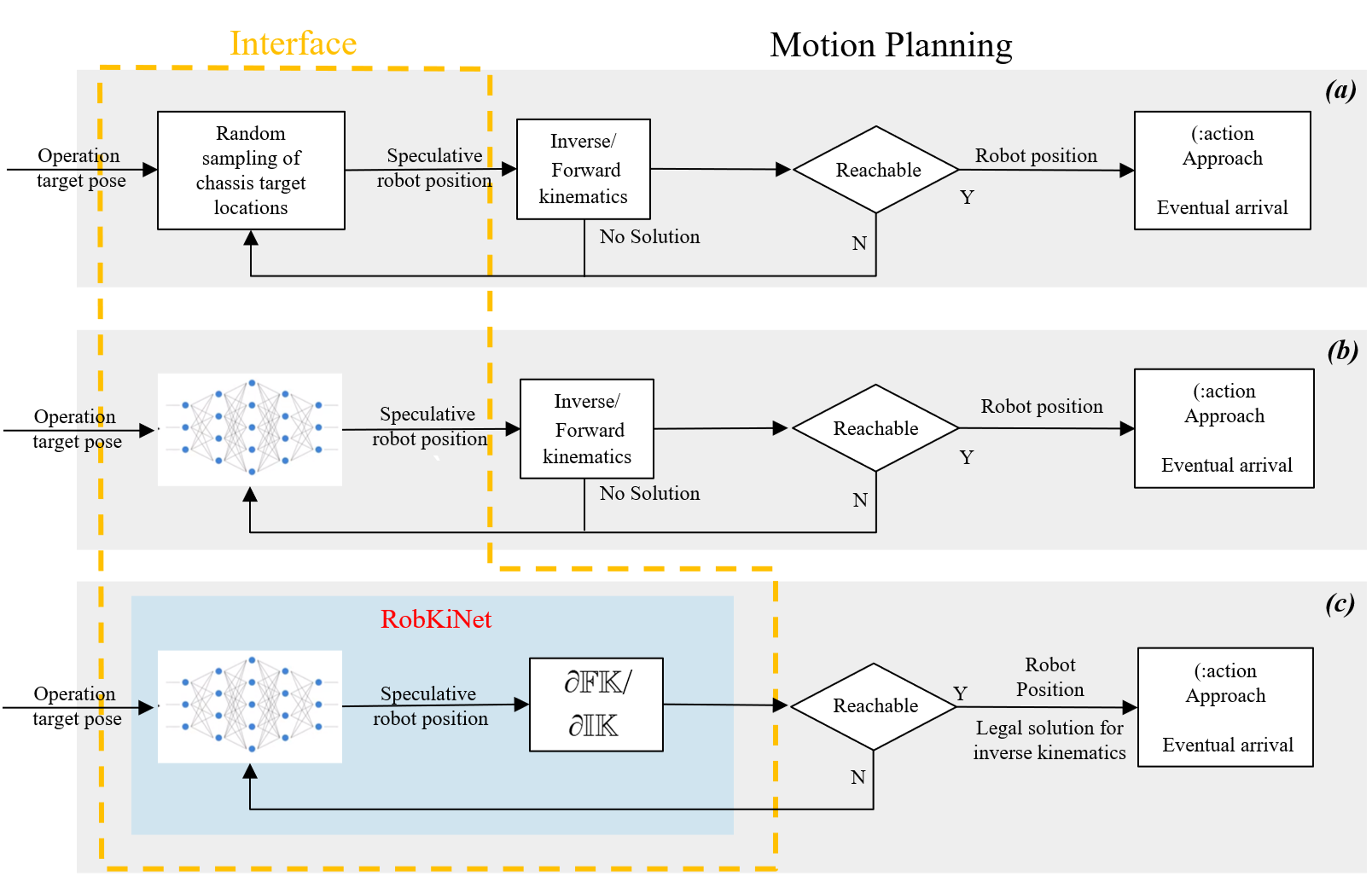}
    \caption{Comparison of three robot configuration (such as the chassis position) sampling methods. (a) Randomized sampling. (b) NN Regression method. (c) RobKiNet proposed in this work. The effectiveness of sampling in a continuous solution space will greatly affect the efficiency of motion planning. RobKiNet eliminates the need to iterate several times to validate the kinematics in a simulation environment.}
    \vspace{-4mm}
    \label{FIG:2}
\end{figure*}

\subsection{RobKiNet Principle: Integrating Neural Networks with Physical Constraints}

Based on the problem definition presented earlier, we will explain why RobKiNet is needed and outline its implementation principles by comparing it with traditional methods.

We often rely on prior knowledge to ensure that a robot's interaction with the environment is safe and reliable. Traditional methods involve setting up precise simulation models tailored to different scenarios and interaction environments. However, this is difficult to achieve in the real world's highly dynamic and unknown environments. The world model approach mentioned earlier typically uses data-driven networks to guide the robot, but data faces the problem of long-tail sparsity \cite{zhou2022dynamically,zhou2018deep}. The constraints represented by datasets are limited, and in unfamiliar scenarios, it is challenging to achieve zero-shot learning or one-shot learning. This may lead to the robot violating physical laws.

To address these issues, our approach is to enable the end-to-end network to learn robotics physical constraints rather than the latent mappings of a dataset. This is the essence of the RobKiNet. 
Our RobKiNet is designed to inject the robot's kinematic constraints as physical constraints into the network to guide the sampling of the robot configuration in the interface between the task and motion planning layers. We will explain the RobKiNet by using the example that injects the differentiable kinematics to solve the sampling problem of infinite solutions in continuous space.

Figure \ref{FIG:2} shows the differences and distinctions between determine the robot configuration using the different pipelines. Figure \ref{FIG:2}(a) is the common method of the PDDL-STREAM framework \cite{pddlstream, PDDLSTREAM2}. For the chassis target position, the RS method presents an assumed chassis position, which is then employed in the robot's inverse kinematics calculation to determine the feasibility of a valid solution and the robot's reachability. However, RS methods often necessitate closed-loop feedback and multiple sampling iterations to determine the correct final position. This prolonged process results in increased task duration and multiple recalculations for the robot's navigation. With the help of neural networks (Figure \ref{FIG:2}(b)), this process can be optimized using a large amount of data training, through which we hope that the Artificial Neural Network(ANN) can learn the implicit mapping relationship between the current position and the target position, thus outputting a more credible result. Due to its supervised learning essence, we call it the NN Regression approach in our experiments.

Compared to the previous two methods, the most significant feature of our RobKiNet is the substantial expansion of the network's end-to-end scope. The network is designed to incorporate physical constraint that we want it to learn, such as complex kinematics. The greatest advantage of neural networks is that they can be expressed as a computational graph, where all forward and backward propagations are executed through the chain rule, making them a series of differentiable mathematical processes. Therefore, the physical constraints we incorporate must be differentiable. 

With the introduction of differentiable kinematics, the training of the network undergoes a fundamental transformation (in Eq. \ref{Differentiable Kinematics}). During forward propagation, we impose penalties and losses on the computational graph nodes that do not satisfy the physical constraints. During backward propagation, the network can then produce optimal parameters through gradient optimization. The RobKiNet process can be exemplified by Figure \ref{FIG:2}(c) which consists of two modules: the Position Prediction Network and the Differentiable Kinematics Calculation Engine. When the pose of the target object (such as the bolt to be disassembled) is inputted, the predictors can directly output the position that the chassis should reach(in CMP and FMP) and provide the corresponding legal solutions for each joint(in FMP), which is approximate to the subconscious intuition of human beings. This method will greatly improve the efficiency of motion planning.

It is worth emphasizing that the fundamental difference between our RobKiNet and traditional neural networks lies in the fact that RobKiNet learns how to adhere to physical constraints during training, rather than learning a specific dataset mapping.
For the predictors, the forward and inverse kinematics algorithms are added to the training process of the neural network in a differentiable way and are involved in learning to steer the output values of the network before backpropagation. The traditional neural network computational graph will be greatly changed due to the addition of the robotic differentiable computation as a computational node after the forward propagation.

\section{Fundamentals of Kinematic Differentiability}\label{fundamental knowledge}

The implementation of Differentiable Kinematics(in Eq. \ref{Differentiable Kinematics}) is achieved by identifying and replacing non-differentiable operators in the kinematics computation with corresponding differentiable operators. For instance, redefining operators at the limits of function values or at discontinuities ensures the completeness of the computational graph. This section will provide a fundamental understanding of kinematics and differentiable programming, conceptually explaining the feasibility and operational logic behind their integration.

\subsection{Foundations of Differentiable Programming}

The primary objective of differentiable programming is to enable automatic differentiation (AD) \cite{bartholomew2000automatic}, a critical tool in modern computational frameworks. AD operates by systematically applying the chain rule of calculus to compute derivatives efficiently, ensuring precise gradient propagation through complex systems. 
The computation graph plays a pivotal role in this process, as it must be designed to support the seamless flow of gradients across its nodes. 

Assume that $\mathcal{O}_{1}, \ldots, \mathcal{O}_{n}$ are a total of n operators that together form a function $f: \mathbb{R}^s \to \mathbb{R}^t$, which completes the mapping from input to output in our system.

The nodes of the computational graph of the mapping represent operators and the edges represent dependencies. An operator $\mathcal{O}_i: \mathbb{R}^{s_i} \to \mathbb{R}^{t_i}$ in the graph is a differentiable operator only if it satisfy the condition that the gradient $\nabla \mathcal{O}_i$ exists everywhere.
Then, the backpropagation can be symbolized as applying the chain rule to the composite function represented by the computational graph:
\begin{equation}\label{operators_propagated}
\nabla f(x) = \nabla \mathcal{O}_n (\mathcal{O}_{n-1}) \cdot \nabla \mathcal{O}_{n-1} (\mathcal{O}_{n-2}) \cdot \ldots \cdot \nabla \mathcal{O}_1(x)
\end{equation}

Each operator $\mathcal{O}_i$ represents the fundamental building block of the computation graph, which can range from a simple linear operation to a larger-scale code block composed of multiple operations. Our goal is to express the kinematics process as a combination of such differentiable operators, thereby enabling full differentiability. This modular approach to designing differentiable programs allows us to control the granularity of automatic differentiation, facilitating an efficient differentiable kinematics process.

\begin{figure*}[t]
    \centering
    \includegraphics[scale=0.17]{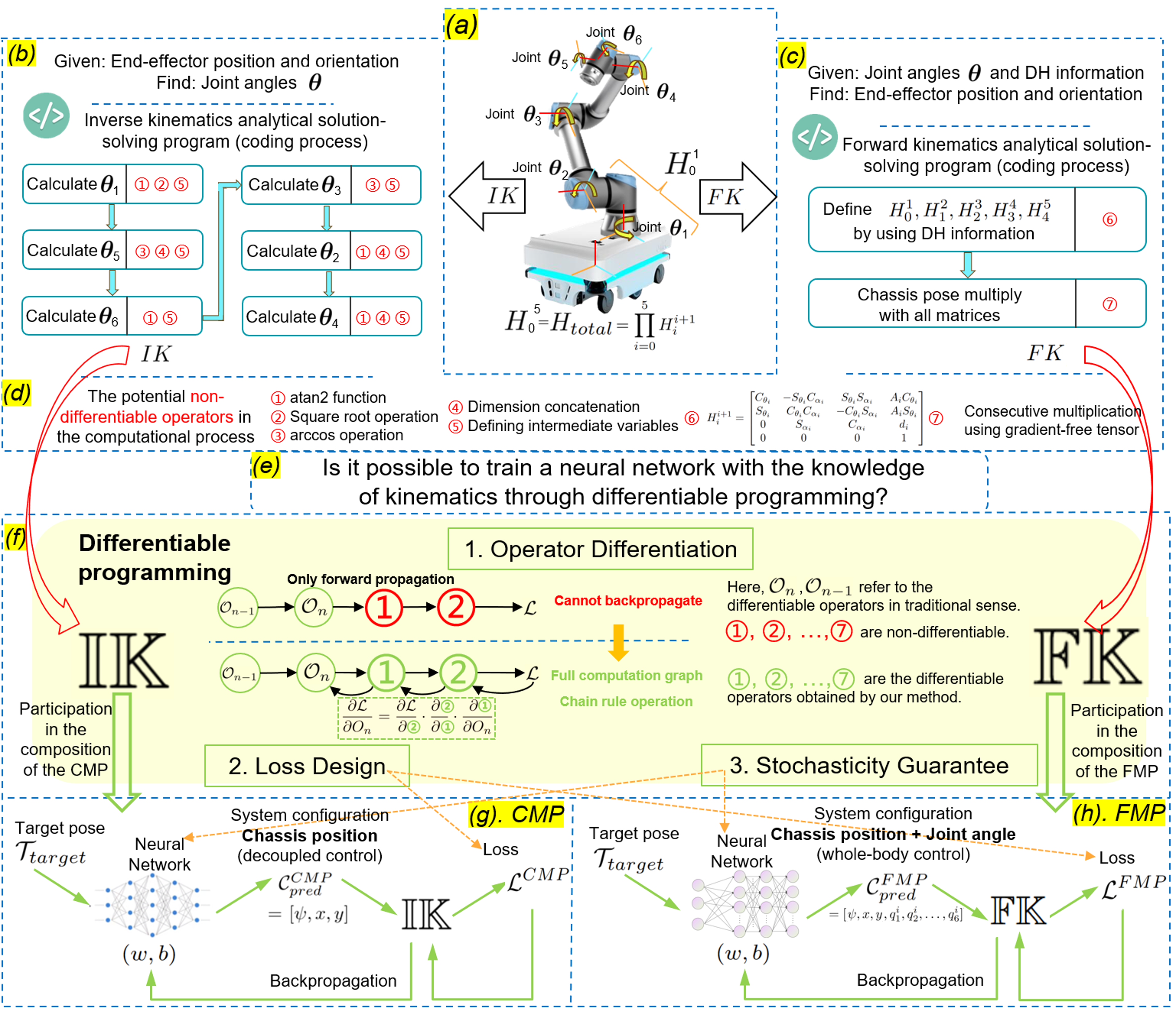}
    \caption{An overview of differentiable kinematics and its integration with neural networks. (a) Overview of the joint coordinate systems for the 6-DOF robotic arm kinematics. (b) The computational process for the analytical inverse kinematics solution. (c) The computational process for the analytical forward kinematics solution. (d) Non-differentiable operations and function modules within the program. (e) The motivation behind this study. (f) The basic concept of differentiable programming — ensuring the chain rule of the computational graph. This is achieved by transforming non-differentiable operators (red circles) into differentiable operators (green). There are other challenges to differential programming (2, 3), and solutions to all these problems will be detailed later. (g) The integration of $\mathbb{IK}$ with neural networks for chassis position prediction (CMP). (h) The integration of $\mathbb{FK}$ with neural networks for predicting both chassis position and joint angle configuration (FMP).}
    \vspace{-3mm}
    \label{Overview of Differentiable Kinematics}
\end{figure*}

\subsection{Differentiable Kinematics Through Differentiable Programming}

In this paper, the kinematics algorithm follows the Denavit-Hartenbarg(DH) convention (Figure \ref{Overview of Differentiable Kinematics}(a)) \cite{DH}, and the homogeneous transformation matrix of any joint $i$ concerning its predecessor joint $i-1$ can be expressed as:
\begin{equation}
H_i^{i-1}=A_i=Rot_{z, \theta_i} \operatorname{Trans}_{z, d_i} \operatorname{Trans}_{x, a_i} Rot_{x, \alpha_i}\label{eq:1}
\end{equation}

Where $\theta$ is the joint angle, $d$ is the linkage offset, $a$ is the linkage length, and $\alpha$ is the linkage twist.
For a manipulator with a known configuration, we can construct a DH table and extract the corresponding hyperparameters for each joint from it. 
Then the forward kinematics formulae for a 6-DOF robotic arm is:
\begin{equation}\label{equation_H_{total}}
H_{total}=\prod_{i=0}^{5}H_i^{i+1} 
\end{equation}

The $H_{total}$ represents the coordinate relationship between the end joint and the base, consisting of rotational attitude and relative position:
\begin{equation}\begin{aligned}
H_{total}=\left(\begin{array}{ll}R & d \\ 0 & 1\end{array}\right), \\ \quad R \in S O(3) , \quad d \in \mathbb{R}^3
\end{aligned}\end{equation}

Figure \ref{Overview of Differentiable Kinematics}(b),(c), and (d) illustrates the analytical solution-solving procedure for the two kinematic constraints involved in our study, as well as the non-differentiable parts encountered in the solution of the program.
When programming the computation for FK(Figure \ref{Overview of Differentiable Kinematics}(c)), we need to define each transformation matrix $H_i^{i-1}$ and use matrix multiplication to obtain the final result. However, this definition and multiplication operations are non-differentiable because they consist of independent scalar elements, and the partial derivatives within these operators cannot be propagated backward to the preceding operator.

The IK calculation works in the opposite way to FK(Figure \ref{Overview of Differentiable Kinematics}(b)). Given the transformation matrix $T_n^0$ between the end-effector and the base, we aim to find the inverse solution by seeking one or more solutions to the following equations:
\begin{equation}\begin{gathered}
T_n^0\left(q_1, \cdots, q_n\right)=H_{total} \\ T_n^0\left(q_1, \cdots, q_n\right)=A_1\left(q_1\right) \cdots A_n\left(q_n\right)
\end{gathered}\end{equation}

The $q_n$ denotes the value of the nth joint, which is exactly the quantity we need to solve for in inverse kinematics. 
Therefore, the process of obtaining analytical solutions for inverse kinematics (IK) is more complex, as the program needs to sequentially solve for the possible values of the six joint angles. This involves several operations that are difficult to differentiate, such as the use of the atan2 function or the definition of independent intermediate variables. These steps can cause the operators in the computational graph to lose their connected edges, resulting in the loss of dependencies.

We list some of the FK and IK processes in (Figure \ref{Overview of Differentiable Kinematics}(d)) as non-differentiable computational processes that are marked in red to represent their inability to participate in the computational graph.
Thus, the core motivation of our research(Figure \ref{Overview of Differentiable Kinematics}(e)) is to seek a unified approach by using differentiable programming technology to differentiate programs in both machine learning and scientific domains.

Our core solution is to individually define each non-differentiable operation, transforming it into a differentiable operator. For instance, as illustrated in Figure \ref{Overview of Differentiable Kinematics}(f), in the chain rule operation process, $\mathcal{O}_{n}$ can refer to any differentiable operator in the calculation. Non-differentiable code processes (red circles 1 to 7) are encapsulated separately, becoming operators that perform the function of the operators(such as the atan2 function) with both forward and backward propagation capabilities (green circles 1 to 7). 
The detailed methodological implementation of this transformation process is described in \ref{methods}, where we summarize several types of representative, generalizable, and generalizable methods.
This allows the gradient to be propagated under the chain rule within the local computational graph(in Eq. \ref{operators_propagated}).
Thus implements differentiable programming $\{IK, FK\} \rightarrow \{ \mathbb{IK}, \mathbb{FK} \}$.

It is important to note that in subsequent computations, the neural network output $\mathcal{C}_{pred}$ typically consists of one-dimensional position and orientation information. This cannot be directly utilized by the differentiable $\mathbb{IK}$ and $\mathbb{FK}$ since kinematics require a homogeneous transformation matrix. We achieve this conversion through a transformation function, which must also be differentiable and become one of the operators.
\begin{equation}\label{Hdefine}
H_{pred}, H_{tar}  = \text{Shaping operator}({C}_{pred},\mathcal{T}_{target})
\end{equation}

We use $H_{pred}$ and $H_{tar}$ to represent the homogeneous transformation matrix form of $\mathcal{C}_{pred}$ and $\mathcal{T}_{target}$, respectively.

Figure \ref{Overview of Differentiable Kinematics}(g) and (h) illustrate the integration of our differentiated kinematic computation engines, $\mathbb{IK}$ and $\mathbb{FK}$, with the neural network, providing a direct response to the research motivation. Based on the form of network outputs used for different controls, we define CMP and FMP under the guidance of the RobKiNet principle. This core methodology will be elaborated in the following Section.

\section{Predictors Training and Differentiable Details}\label{RPSN Training Based on Differentiable Arithmetic Process Guidance}

In this section, we will discuss how we utilize RobKiNet to create two separate predictors to tackle the sampling issue of configuration in continuous solution spaces. We have chosen two fundamentally different kinematic constraints to execute distinct control strategies. Additionally, we will provide details on the implementation of differentiable programming.

\subsection{Computational Graph Construction and the CMP Training for Decoupled Control}
We first define CMP using our RobKiNet in conjunction with more complex inverse kinematics(Figure \ref{Overview of Differentiable Kinematics}(g)).
The structure of the CMP network requires that its robotic inverse kinematics computational engine be written into the training process in a differentiable form. That means neither existing computational functions nor third-party libraries can be added as nodes to the computational graph of the neural network backpropagation due to their discontinuities.

For the neural network input tensor, as in the previous example of the bolt pose to be disassembled, we give $\mathcal{T}_{target} =[\varphi,\theta,\psi,x,y,z], \mathcal{T}_{target} \in \mathbb{R}^6$, where $\varphi,\theta,\psi$ represent the spatial orientation described using Euler angles. Considering that the robot's actual working process is unchanged compared to the chassis position, its pitch angle, lateral inclination, and longitudinal displacement should be zero. Under the action of the position prediction network, the robot target chassis position can be predicted as:
\begin{equation}
\mathcal{C} _{pred}^{CMP} = \text{Neural Network1}(\mathcal{T}_{target}) = [\psi,x,y] \in \mathcal{C}_{base}
\end{equation}

Then the solution of its inverse kinematics under the differentiable $\mathbb{IK}$ computational engine proposed in this paper can be expressed as:
\begin{equation}
\begin{aligned}
\left\{\begin{matrix} 
  solu_{angle}=\mathbb{IK}(H_{total})  \\H_{total} =((H_{pred})^{Transpose} \cdot H_{tar})
\end{matrix}\right.
\\solu_{angle}\in \mathbb{R}^{8×6}
\end{aligned}
\end{equation}

Where $H_{pred}$ and $H_{tar}$ are defined in Eq. \ref{Hdefine} by transforming the neural network's output $\mathcal{C} _{pred}^{CMP}$ into Eulerian coordinates, followed by conversion into a rotation matrix, and ultimately into a homogeneous transformation matrix. The $\mathbb{IK}$ is the differentiable computational engine for the inverse kinematics algorithm. The inverse kinematics solves eight sets of solutions, each with six joint values, corresponding to the six joints of the six-axis robotic arm, which can be expressed as Eq. \ref{eq:66}.

During programming, to make this solution process differentiable, we redefine the traditional matrixed kinematics solution by Figure \ref{Overview of Differentiable Kinematics}(f). This ensures that the robot's kinematics solution can be added to the computational graph. 
The introduction of differentiable computational processes like $\mathbb{IK}$ into the computational graph significantly increases its complexity.   
The computational graphs illustrating actual participation in CMP and the following FMP are available for download and inspection at the following website\footnote{\url{https://sites.google.com/view/sjtu-robkinet}\label{fn:supplementary}}.


\subsection{FMP Training for Whole-body Control}

Most of the mobile manipulators currently in use employ a loosely coupled integration between the chassis and the robotic arm, with each being controlled independently by separate controllers. The design of the CMP specifically addresses this loosely coupled scenario, where the network outputs only the parameters for the chassis. However, a more advanced approach involves whole-body control \cite{minniti2019whole, li2020design, zhang2023end}, enabling coordinated control of the entire system. Consequently, we aim for the predictor to simultaneously provide both the sampled positions of the chassis and the six joint angles just like the expanded DH method \cite{zhang2023end}. This prediction, made without any actual physical action, must still accurately adhere to the physical constraints of forward kinematics.

To achieve this, we adopted the RobKiNet and selected the second type of kinematic constraint, constructing the FMP specifically for whole-body control. FMP can directly output the chassis position along with the anticipated posture at that position, aligning more closely with human intuition and the subconscious behavior we exhibit when interacting with the world.

The principle of the FMP can be expressed by the following equation. With the input target pose $\mathcal{T}_{target}$, we predict the whole-body control system parameters:
\begin{equation}
\mathcal{C} _{pred}^{FMP} = \text{Neural Network2}(\mathcal{T}_{target})
\end{equation}

\begin{equation}\label{FMP9output}
    \mathcal{C}_{pred}^{FMP} = 
  [\underbrace{\psi,x,y}_{^{P}\mathcal{C}_{pred}^{FMP}}, \underbrace{q_1^{i}, q_2^{i}, \dots, q_6^{i}}_{^{R}\mathcal{C}_{pred}^{FMP}}]^T,\mathcal{C}_{pred}^{FMP}  \in \mathbb{R}^9
\end{equation}

According to Figure \ref{Overview of Differentiable Kinematics}(h), we need to use differentiable forward kinematics $\mathbb{FK}$ to calculate the prediction error:
\begin{equation}\label{FMP_error}
    \mathbb{FK}(^{R}\mathcal{C}_{pred}^{FMP},\hspace{0.05cm}^{P}\mathcal{C}_{pred}^{FMP}) = H_{tar}  \pm \Delta d
\end{equation}

Here, the forward kinematics formulae derived from Eq. \ref{eq:1} for a six-axis robotic arm are presented in Eq. \ref{equation_H_{total}}. 
$\mathcal{C}_{pred}^{FMP}$ is the output of the FMP for whole-body control, which consists of the sample for the chassis $^{P}\mathcal{C}_{pred}^{FMP} \in \mathcal{C}_{base}$ and the corresponding joints $^{R}\mathcal{C}_{pred}^{FMP} \in \mathcal{C}_{arm}$ for the manipulator. 
$\mathbb{FK}$ is the differential calculus form of Eq. \ref{equation_H_{total}}.
$\Delta d$ is the prediction error. Essentially, this involves forward-propagating the predicted parameters through the computational graph to assess accuracy before any action is taken. In FMP we allow 
$\Delta d \leq 1mm$.
This means that no algebraic solvers are required, the positional error between the robotic arm and the actual target remains within a 1mm range based on the network's predicted configuration.

\subsection{Details of the differentiable methodology}\label{methods}

In the process of differentiating the entire computation, we present the following three difficult problems (shown in Figure \ref{Overview of Differentiable Kinematics}(f)):

\begin{enumerate}[(i)]
    \item  Not all parts of forward and inverse kinematics can be easily differentiated. How can discontinuous operators be differentiable and participate in the computational graph?
    \item 	CMP and FMP should learn how to follow physical constraints during training. How can the loss function be defined to effectively guide the training process?
    \item  The parameters of the differentiable network are fixed after the training is completed. How to ensure its stochasticity to make the ideal solution space large enough?
\end{enumerate}
\subsubsection{Operator Differentiation Methods}

\begin{figure*}[t]
    \centering
    \includegraphics[scale=0.17]{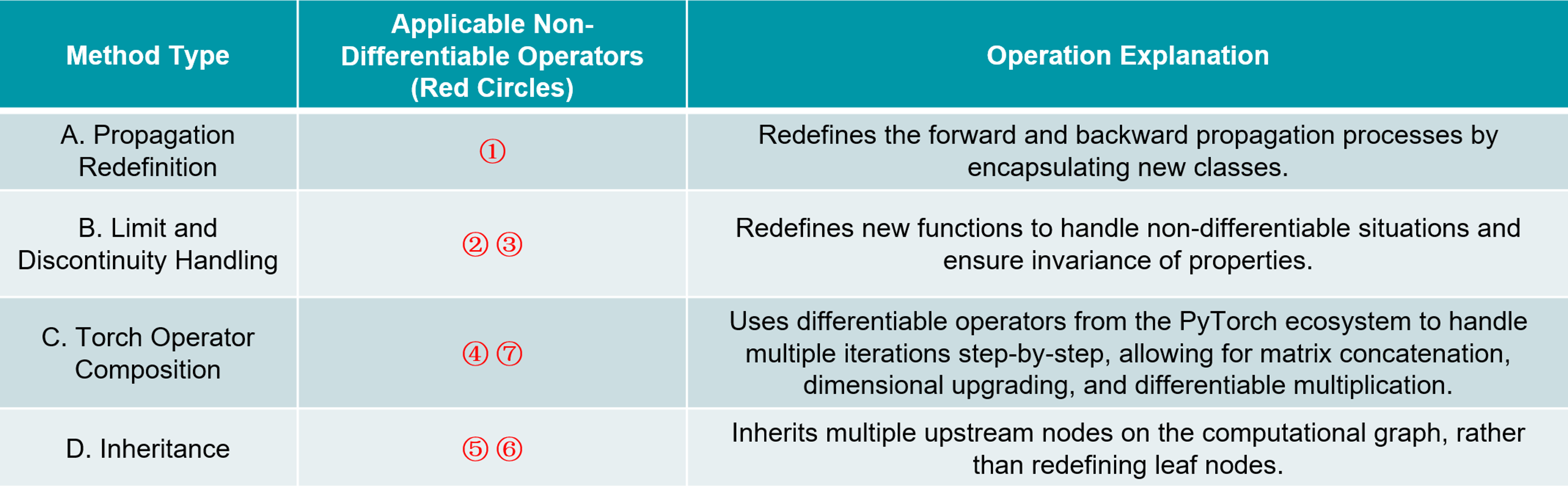}
    \caption{Differential Method summary, corresponding to the treatment of non-differentiable operators in differential kinematics(red circle in Figure \ref{Overview of Differentiable Kinematics}).}
    \vspace{-3mm}
    \label{Differential Method.png}
\end{figure*}

For the first question, we have compiled a list of commonly used differential methods in differentiable programming (Figure \ref{Differential Method.png}). These methods assist us in handling various non-differentiable operators. 

\noindent \textbf{Core Idea: First, identifying the inputs and outputs of the operator; Then, ensuring they originate from or point to other nodes within the computation graph; Finally, maintaining inner continuity without leaf nodes.}

Methods C and D in the figure are relatively straightforward, as the PyTorch framework already provides several computational operators useful for algorithm development, including but not limited to torch.mm for matrix multiplication and torch.stack for dimensional reorganization. We simply need to follow the Core Idea in our programming.
We will illustrate the general methodology with examples of the atan2 function (method type A) and the $arccos$ function (method type B) during CMP training.

A widely adopted practice in robot inverse kinematics involves leveraging the atan2 function as an efficient solution for calculating inverse tangents.
However, the standard form of the atan2 function is discontinuous, causing direct interruptions in gradient flow within the computational graph. Hence, it becomes imperative to redefine both the forward and backward propagation processes of the function itself, as illustrated in Algorithm \ref{alg:1}.
\begin{algorithm}
\caption{Differentiable Four-Quadrant Arctangent }\label{alg:1}
\begin{algorithmic}[1]
\State \textbf{class} Atan2Function:\Comment{From torch. autograd.Function}
\quad\quad\Function{forward}{$y, x$}
\State $result \gets$ \Call{calculate\_atan2}{$y, x$}
\State $ctx  \gets$\Call{save\_for\_backward}{$x, y$}
\State  \Return \Call{tensor}{$result, \text{requires\_grad} = \text{True}$}  
\quad\quad\EndFunction
\quad\quad\Function{backward}{$\text{ctx}, \text{grad\_output}$}
\State \quad $x, y \gets$ \Call{saved\_tensors}{$\text{ctx}$}
\State \quad$grad\_y \gets$$\frac{\mathrm{d}}{\mathrm{d}y}\text{atan2}(y, x)\gets x / (x^2 + y^2)$
\State \quad$grad\_x \gets$$\frac{\mathrm{d}}{\mathrm{d}x}\text{atan2}(y, x)\gets -y / (x^2 + y^2)$
\State \quad \Return $\text{grad\_output} \times grad\_y, \text{grad\_output} \times grad\_x$
\quad\quad\EndFunction
\Function{main}{}
\State \quad $y \gets$ \Call{get\_input\_y}{}  
\State \quad $x \gets$ \Call{get\_input\_x}{} 
\State \quad $atan2 \gets$ \Call{Atan2Function.apply}{$y, x$}
\EndFunction
\end{algorithmic}
\end{algorithm}

Atan2Function defines its propagation process in the form of a class. While forward propagation computes the results and stores the values of the nodes, the derivation of backpropagation can be expressed as the gradient of the node equals the result of multiplying the partial derivatives of that node by the cumulative gradient passed from the previous node. The relationship can also be seen as a concrete application of the chain rule. In this way, this inherently non-differentiable function successfully participates in the computational graph. 
It turns out that by simply changing the function's body, this approach can be applied to address a range of formally discontinuous functions, ensuring computational differentiability. 

Another frequent differentiable error in inverse kinematics occurs when an intermediate variable falls outside the domain of the mathematical function that operates on it. When confronted with this problem, the core idea is to continue to redefine the propagation operations of the function. Taking the $arccos$ function as an example,
when it appears that the value of the previous node in the computational graph is outside its domain of definition, we construct a new customized extended differentiable function $G(x)$, with the core arithmetic function $arccos$ of the current node as a base function. Taking \(\delta > 0\) and \(\delta\) as sufficiently small:
\begin{equation}
\setlength{\abovedisplayskip}{3pt}
\setlength{\belowdisplayskip}{3pt}
G(x) = \begin{cases}
  -\frac{(x-(-1+ \delta))}{\sqrt{1-(-1+ \delta)^2}} + \cos^{-1}(-1 + \delta),  \text{if } x \leq -1 + \delta, \\
  \cos^{-1}(x), \hspace{1.8cm}\text{if } -1 + \delta< x < 1 - \delta, \\
  -\frac{(x-(1- \delta))}{\sqrt{1-(1- \delta)^2}} + \cos^{-1}(1 - \delta),  \text{if } x \geq 1 - \delta.\label{eq:9}
\end{cases}\end{equation}

In our formula, we employ the concept of tangent lines for construction. 
This adjustment is made to ensure that within the domain of the $arccos$ function definition, it operates normally and contributes to defining the loss function. At the same time, the unsatisfactory part outside the domain of definition can have such a large slope that it is penalized by the network preference without interrupting the computation of the graph. 
For this purpose, it is imperative to guarantee that the newly constructed functions exhibit identical values of the first-order derivatives at the points of discontinuity.

This ensures that even if the value passed in from the previous node is out of the definition domain, the whole computation engine will not be terminated due to error reporting. In practice, this change not only ensures the program's normal execution but also allows us to incorporate the out-of-definition domain into the loss composition. This is beneficial because this region typically has a steeper slope, resulting in a larger contribution. During backpropagation, the out-of-definition domain is penalized preferentially, represented as $L^{outdom}_k$ in Eq. \ref{eq:outdom}.

\subsubsection{Training Core: Unique Definition of Loss}

The solution to the second difficulty is inspired by the fact that, when we do not rely on the loss defined by the difference between the output and labels, the only pointers to training come from the various types of anomalies and their defined losses during the computation. It is crucial not only to inform the network whether its outputs comply with kinematic constraints but also to effectively communicate the degree of deviation from those constraints.

\noindent \textbf{Core Idea: Penalize all differential operators that may yield abnormal results or fall outside the ideal solution space. Losses should reflect deviations or degrees of abnormality.}

If we integrate these losses into the guidance of the neural network, it will generate outputs capable of traversing all computational pathways, thereby culminating in a valid solution. Naturally, this is a qualitative analysis, while the quantitative calculation of the predictors' losses is provided by the following equations. For clarity, the various losses, their corresponding meanings, and the informed prior knowledge are summarized in Table \ref{tab:loss functions}.

\begin{table*}[ht]
\centering
\caption{Meaning of Customized Loss and Their Corresponding Informed Prior Knowledge} \label{tab:loss functions}
\renewcommand{\arraystretch}{1.5} 
\begin{tabular}{|c|>{\centering\arraybackslash}m{0.5\textwidth}|c|}
\hline
\textbf{Loss Components} & \textbf{Meaning and Function} & \textbf{Informed Prior Knowledge}\\
\hline
$\mathcal L^{illroot}_k$ & Illegal Root Loss: Penalizes when a node in the computation graph results in a  negative value under a square root.  & Non-negative values under the square root.\\
\hline
$\mathcal L^{outdom}_k$ & Out of Domain Loss: Penalizes when intermediate variables exceed the operator's defined domain.  & Domain constraints on independent variables.\\
\hline
$\mathcal L^{illsolu}_k$ & Illegal Solution Loss: Penalizes when the inverse kinematics has  solutions, but the joint angles contain illegal values.  & Joint angle limits.\\
\hline
$\mathcal L^{idesolu}_k$ & Ideal Solution Loss: Do not penalizes when at least one valid solution for inverse kinematics exists.  & At least one valid solution.\\
\hline
$\mathcal L^{pre\_error}_k$ & Predicted Result Error Loss: Penalizes the prediction error after forward kinematics of the FMP predicted configuration.  & High-precision fitting within 1mm.\\
\hline
$\mathcal L^{distance}_k$ & Distance Error Loss: Penalizes errors in the appropriateness of the predicted chassis position range by the FMP.  & Goal within the workspace.\\
\hline
$\mathcal L^{orien}_k$ & Orientation Error Loss: Penalizes errors in the end-effector's orientation under the FMP predicted joint configuration.  & Maximum allowable orientation angle of 0.1 rad.\\
\hline
\end{tabular}
\end{table*}

\noindent \textbf{For CMP:} 
\begin{equation}
{
\begin{aligned}
\mathcal L^{CMP}=\frac{1}{N} \sum_{k=1}^N (&\mathbb U \cdot  \delta_{illroot}  \mathcal L^{illroot}_k+\mathbb U \cdot \delta_{outdom} \mathcal L^{outdom}_k \\ +  &\delta_{illsolu} \mathcal L^{illsolu}_k + \delta_{idesolu} \mathcal L^{idesolu}_k\label{eq:10}
 )
\end{aligned}
}
\end{equation}

Where $\delta$ represents the weighting relationship between the different parts of the losses, which will determine which losses are preferentially penalized during backpropagation.
The $\mathcal L^{illroot}_k$ as well as $\mathcal L^{outdom}_k$ denote exactly the penalties for the location prediction network data on the wrong branch during the differentiable computation process. They guarantee that the inverse operation has the result as shown in Eq. \ref{eq:6}, even if the solved angle value is illegal. Denote certain intermediate variables in the computation of $q_{j}^{i}$ as the symbol $q^{pre}$, the specific formula definitions are as follows:

To ensure that the $f\left(q^{pre}\right)=\sqrt{q^{pre}}$ is legalised:
\begin{equation}
{
\mathcal L_{k}^{illroot}=\left\{\begin{array}{ll}
0, & \text { if } q^{pre} \geqslant 0 \\
-q^{pre}, & \text { if } q^{pre}<0
\end{array}\quad \right.
}
\end{equation}

To ensure that the $G\left(q^{pre}\right)$ in Eq. \ref{eq:9} is legalised:
\begin{equation}
{
\mathcal L_{k}^{\text {outdom}}=\left\{\begin{array}{cl}
0, & \text { if } q^{pre} \in[-1, 1] \\
|q^{pre}|-1, & \text { if } q^{pre} \notin[-1, 1]\label{eq:outdom}
\end{array} \quad 
\right.
}
\end{equation}

For numerically illegal angle values beyond $ [-\pi, \pi]$, we use $\mathcal L^{illsolu}_k$ to penalize them. Obviously, if there exists any one of the eight sets of solutions where all the angular values of the angular solutions are within the legal range, the loss will be defined as 0, that is, $\mathcal L^{idesolu}_k$. The specific formula definitions are as follows:
\begin{equation}
{
\mathcal L_{k}^{\text {illsolu}}=\left\{\begin{array}{cl}
0, & \text { if } q_{j}^{i} \in[-\pi , \pi] \\
|q_{j}^{i}|-\pi, & \text { if } q_{j}^{i} \notin [-\pi , \pi]
\end{array} \quad 
\right.
}
\end{equation}

\begin{equation}
{
\mathcal L_{k}^{\text {idesolu}}=\left\{\begin{array}{cl}
0, & \text { if } \exists i \in [1,8], \forall j \in [1 , 6], q_{j}^{i} \in[-\pi , \pi] \\
L_{k}^{\text {illsolu}}
\end{array} \quad 
\right.
}
\end{equation}

In particular, for the inverse kinematics computation process included in the CMP, we specifically define $\mathbb U$ in Eq. \ref{eq:10}. When it takes 1, the loss function takes into account the loss from all the errors in the inverse kinematics computation process that deviates from the correct data flow, which is the task-specific loss. When it takes 0, we only define the loss and guide the training for one or more sets of illegal values in the final angular solution. Apparently, the former is more relevant for the present application scenario and the loss is defined more strictly, while the latter is an attempt of ours regarding the definition of loss in general problems and generalized scenarios.

\noindent \textbf{For FMP:} 
\begin{equation}
{
\begin{aligned}
  \mathcal L^{FMP}=\frac{1}{N} \sum_{k=1}^N ( \delta_{pre\_error}  \mathcal L^{pre\_error}_k + & \delta_{distance} \mathcal L^{distance}_k\\ + & \delta_{orien} L^{orien}_k  \label{For RPSN-FK}
 )
\end{aligned}
}
\end{equation}

\begin{equation}
\begin{aligned}
{
\mathcal L_k^{\text{pre\_error}}=\left\{\begin{array}{cl}
\text{MSELoss}(\mathbb{FK}(^{R}\mathcal{C}_{pred}^{FMP}, & ^{P}\mathcal{C}_{pred}^{FMP}) - H_{\text{tar}}),\\ &\text{ if } \Delta d \geq \text{ 1 mm} \\
0, & \text { if } \Delta d < \text{ 1 mm}
\end{array} \quad 
\right.
}
\end{aligned}
\end{equation}

\begin{equation}
{
\mathcal L_k^{\text{distance}}=\left\{\begin{array}{cl}
\text{MSELoss}(^{P}\mathcal{C}_{pred}^{FMP}, \mathcal{T}_{target}), & \text{ if } \mathcal{T}_{target} \notin \mathcal{W} \\
0, & \text { if } \mathcal{T}_{target} \in \mathcal{W}
\end{array} \quad 
\right.
}
\end{equation}

\begin{equation}
\mathcal{L}^{\text{orien}}_k =
\begin{cases}
\sum_{i=x,y,z} \left| \arctan2\left(\sin(\Delta \theta_i), \cos(\Delta \theta_i)\right) \right|, \\  \hspace{4.5cm} \text{if } |\Delta \theta_i| \geq 0.1 \, \text{rad} \\
0, \hspace{4.18cm} \text{if } |\Delta \theta_i| < 0.1 \, \text{rad}
\end{cases}
\end{equation}

Where the loss function is defined by the weighting coefficients $\delta$ and three sub-losses. For FMP, we allow the nine parameters output for whole-body control to have a distance of less than $\Delta d$ from the target position $H_{\text{tar}}$ after being constrained by forward kinematics. 

The degree of deviation is controlled by $\mathcal L_k^{pre\_error}$. $\mathcal L_k^{distance}$ and $\mathcal L^{orien}_k$ impose constraints on the chassis position and the end-effector's orientation, respectively. Here, $\Delta \theta_i$ represents the error between the end-effector's orientation, calculated through forward kinematics, and the target object's orientation in the x, y, and z directions.

\subsubsection{Stochasticity Guarantees}
To tackle the third challenge, an additional Dropout layer was incorporated into the position prediction network of the RobKiNet, as illustrated in Table \ref{table:1}. 

\noindent \textbf{Core Idea: Introduce stochasticity into the forward propagation process without affecting the distribution of sampled results within the ideal solution space, $\mathcal{C}^*$.}

It is worth emphasizing that in existing deep learning networks, Dropout layers prune neurons randomly during training to prevent overfitting. However, the Dropout layers introduced here inject minimal stochasticity during testing, and they do not significantly impact the network's accuracy, as shown in the next section. This approach is inspired by the principle of introducing stochasticity through the temperature parameter \cite{LLM-temperature}, commonly used in large language models.

During testing, the Dropout layer remains active. When the differentiable kinematics calculation engine detects unsatisfactory outputs from the trained model, the testing process is restarted, and the RobKiNet generates a fresh set of solutions. Since no loss is computed and no backpropagation is performed during this phase, the Dropout layer plays a crucial role in introducing stochasticity. Each regenerated result is generated by accessing the well-trained model, and any prior failures do not influence the current outcome.
Stochasticity and accuracy are often inversely related. The principle for selecting the level of stochasticity is that the stochasticity introduced during both training and testing should not be excessive. It must ensure that the sampled solutions fluctuate within our ideal solution space $\mathcal{C}^*$.

\section{Experiments and Results}\label{Experiments and Results}
\subsection{Experiments Setup}


	
	 
	 
 


In order to validate the effectiveness of the proposed method in this work, we use the disassembly of EOL-EVBs bolts as a practical application scenario to explore the performance demonstrated by the CMP and the FMP in terms of training and testing.

\begin{table}[h]
\caption{Training Information for Each Model}
\vspace{-3mm}
\begin{center}
\begin{tabular}{|c|cll|c|c|}
\hline
Model & \multicolumn{3}{c|}{Composition}                                                                                                                               & Loss function                                                                                       & \begin{tabular}[c]{@{}c@{}}Hyperparameter \\ optimization\end{tabular} \\ \hline
{\begin{tabular}[c]{@{}c@{}}NN \\Regression\end{tabular}}   & \multicolumn{3}{c|}{\begin{tabular}[c]{@{}c@{}}input\_layer = 6\\hidden\_layer = 50\\ Dropout\\hidden\_layer = 50\\ output\_layer = 6\end{tabular}}                                     & MSE                                                                                                 & Ray.tune                                                               \\ \hline
CMP1 & \multicolumn{3}{c|}{\begin{tabular}[c]{@{}c@{}}input\_layer = 6\\hidden\_layer = 50\\ Dropout\\hidden\_layer = 50\\ output\_layer = 3\\ $\mathbb{IK}$\end{tabular}} & \begin{tabular}[c]{@{}c@{}}$\mathcal L^{CMP}$\\ $\mathbb{U} = 1$ \end{tabular} & Ray.tune                                                               \\ \hline
CMP2 & \multicolumn{3}{c|}{\begin{tabular}[c]{@{}c@{}}input\_layer = 6\\hidden\_layer = 50\\ Dropout\\hidden\_layer = 50\\ output\_layer = 3\\  $\mathbb{IK}$\end{tabular}} & \begin{tabular}[c]{@{}c@{}}$\mathcal L^{CMP}$\\  $\mathbb{U} = 0$\end{tabular} & Ray.tune                                                               \\ \hline
FMP & \multicolumn{3}{c|}{\begin{tabular}[c]{@{}c@{}}input\_layer = 6\\hidden\_layer = 50\\ Dropout\\hidden\_layer = 50\\ output\_layer = 9\\  $\mathbb{IK}$\end{tabular}} & \begin{tabular}[c]{@{}c@{}}$\mathcal L^{FMP}$ \end{tabular} & Ray.tune                                                               \\ \hline
\end{tabular}\label{table:1}
\vspace{-2mm}
\end{center}
\end{table}
We conducted comparative experiments to demonstrate the difference in training outcomes when incorporating the computational engine for differentiable kinematics into the computational graph compared to a single ANN(the NN Regression method). The specific experimental training information is shown in Table \ref{table:1}. We built the network model based on PyTorch and provided the training of NN Regression with the end-effector position required for the disassembly task in the same coordinate system with a chassis position that can accomplish the disassembly task. The loss between the network output and the ground truth is defined by MSEloss. For the training of the predictors, the input data is a single end-effector pose required for the disassembly task, and its loss can be categorized into different cases according to the definition of Eq. \ref{eq:10} and Eq. \ref{For RPSN-FK}. CMP1 represents a more stringent task-specific loss, while CMP2 represents a more general loss based on the use of a differentiable method as described in the previous section.

To evaluate model performance under varying data volumes effectively, we designed twenty different data volume sets, ranging from a minimum of 20 groups to a maximum of 3200 groups, for training and testing the three models. To ensure that the training hyperparameters are optimal under different data volumes and model conditions, we use Ray-tune \cite{tune} to search for optimal hyperparameters. It is guaranteed that each network model is randomly selected at least 30 times for hyperparameters under each data volume, and the network models trained with optimal hyperparameters are recorded with their test results. 


\begin{table*}[t]
\caption{Experimental Setup Introducing Stochasticity}
\vspace{-4mm}
\begin{center}
\renewcommand{\arraystretch}{1.5} 
\begin{tabular}{|c|c|c|c|c|}
\hline
      & Dropout(Training) & Dropout (Testing) & Epoches & Dropout Rate                         \\ \hline
Case1 & Deactivated       & Deactivated       & 400     & None                                 \\ \hline
Case2 & Activated         & Activated         & 1000    & 2\%-15\%(use Ray-tune)               \\ \hline
Case3 & Deactivated       & Activated         & 400     & Consistent with case2 optimal result \\ \hline
\end{tabular}\label{table:3}
\vspace{-4mm}
\end{center}
\end{table*}

\begin{table*}[h]
\caption{CMP Performance: Sampling Times and Time Costs for Each Method}
\vspace{-4mm}
\begin{center}
\renewcommand{\arraystretch}{1.5} 
\begin{tabular}{|c<{\centering}|c<{\centering}|c<{\centering}|c<{\centering}|c<{\centering}|c<{\centering}|}
\hline
      & \begin{tabular}[c]{@{}c@{}}Average samples\\ (times)\end{tabular} & \begin{tabular}[c]{@{}c@{}}Single sampling time\\ (ms)\end{tabular} & \begin{tabular}[c]{@{}c@{}}Single kinematic\\ simulation (ms)\end{tabular} & \begin{tabular}[c]{@{}c@{}}Total time for one\\  speculation(ms)\end{tabular} & \begin{tabular}[c]{@{}c@{}}Time-reduction\\  factor\end{tabular} \\ \hline
RS    & 31.04                                                             & 0.296                                                               & \multirow{6}{*}{174.48}                                                    & 5425.05                                                                       & 1 $\times$(base)                                                            \\ \cline{1-3} \cline{5-6} 
EBS & 5.80                                                              & 0.473                                                               &                                                                            & 1014.73                                                                        & 5.34 $\times$                                                               \\ \cline{1-3} \cline{5-6}
NN Regression   & 3.06                                                              & 0.315                                                               &                                                                            & 534.87                                                                        & 10.14 $\times$                                                              \\ \cline{1-3} \cline{5-6} 
CMP1 & 1.28                                                              & 0.339                                                               &                                                                            & 223.77                                                                        & 24.24 $\times$                                                             \\ \cline{1-3} \cline{5-6} 
CMP2 & 1.42                                                              & 0.312                                                               &                                                                            & 248.20                                                                        & 21.86 $\times$                                                               \\ \cline{1-3} \cline{5-6}
                                                         \hline
\end{tabular}\label{table:4}
\end{center}
\vspace{-3mm}
\end{table*}

\begin{table*}[h]
\caption{FMP Performance: Solving Time and Analysis}
\vspace{-4mm}
\begin{center}
\renewcommand{\arraystretch}{1.5} 
\begin{tabular}{|c|c|c|c|}
\hline
            & \begin{tabular}[c]{@{}c@{}}Total time for one\\  speculation(ms)\end{tabular} & \begin{tabular}[c]{@{}c@{}}Time-reduction\\  factor\end{tabular} & \begin{tabular}[c]{@{}c@{}}Analysis\end{tabular} \\ \hline
            
RS           & \begin{tabular}[c]{@{}c@{}}30000(meet the maximum \\ and got none result) \end{tabular}       & 1 $\times$(base)       & \begin{tabular}[c]{@{}c@{}}The solution space has too many dimensions, leading to no solution \\ within the given time limit.  \end{tabular}                                     \\ \hline
RS + Jacobian  & 6732.77         & 4.46 $\times$         & Gradient-based methods guide the search but take longer.                   \\ \hline
EBS + Jacobian & 2455.06       & 12.22 $\times$         &     The iteration is more efficient but yields only a single solution.  \\ \hline
DDPG & 311.08       & 96.44 $\times$         &     Excessive data magnitude and high training costs(Figure \ref{FIG:FMP}).  \\ \hline
FMP          & 195.76         & 153 $\times$         & \begin{tabular}[c]{@{}c@{}}Accurately and efficiently provides system configurations to \\ achieve whole-body control.  \end{tabular}                   \\ \hline
\end{tabular}\label{FMP-Jacobian}
\end{center}
\vspace{-3mm}
\end{table*}

\subsection{Experiments Details}
We consider a sampling point to be valid when it satisfies the robot's kinematic constraints, is suitable for the workspace, and does not lead to robot singularities. 
The ideal solution and accuracy have already been clearly defined in Section \ref{Notation and Definitions}.

\begin{figure}[t]
    \centering
    \vspace{-2mm}
    \includegraphics[scale=0.17]{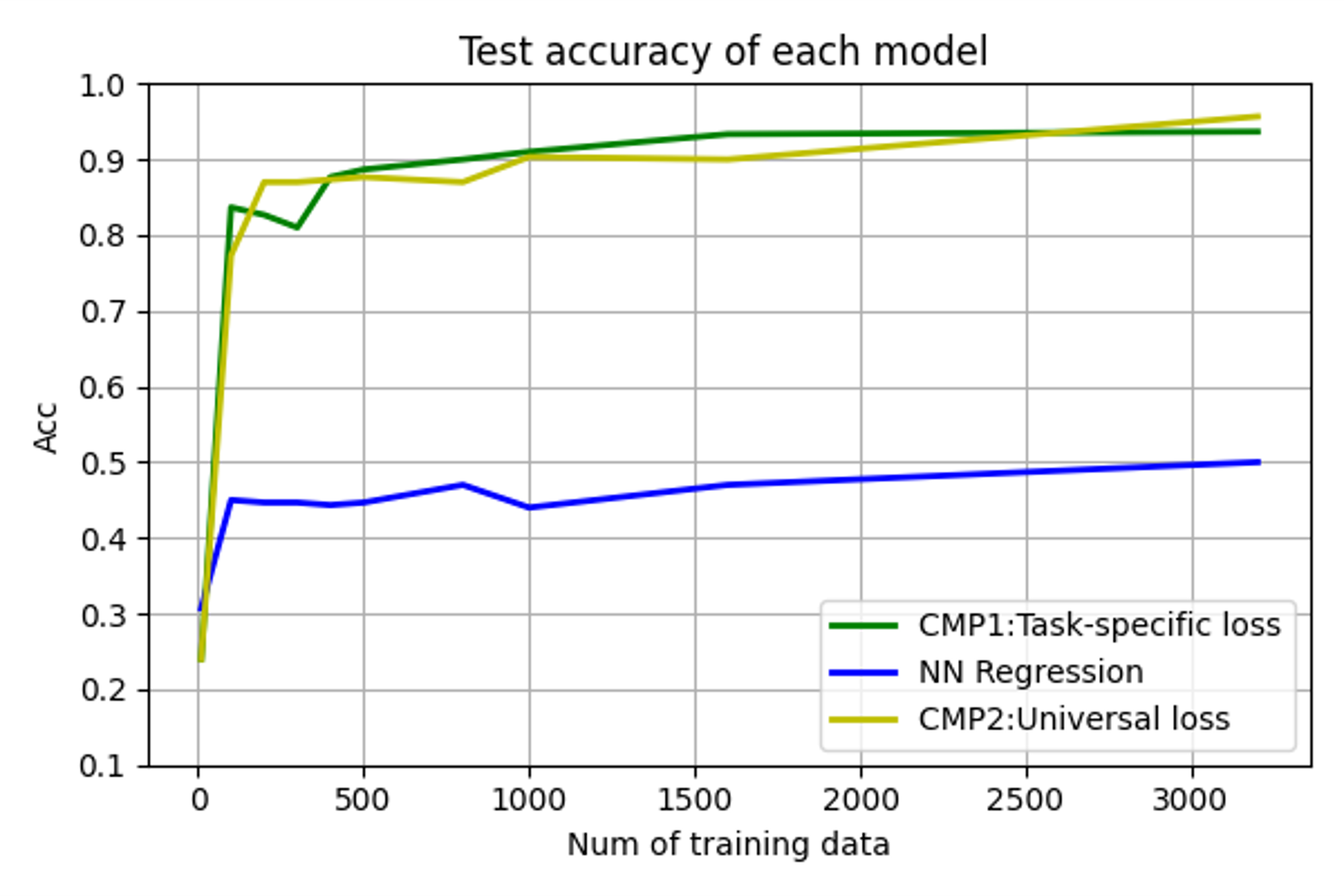}
    \caption{For decoupled control: Test accuracy after training with different data amounts. The data volume ranges from 50 to 3200 groups.}
    \label{FIG:6}
    \vspace{-2mm}
\end{figure}

\begin{figure}[h]
    \centering
    \vspace{-2mm}
    \includegraphics[scale=0.10]{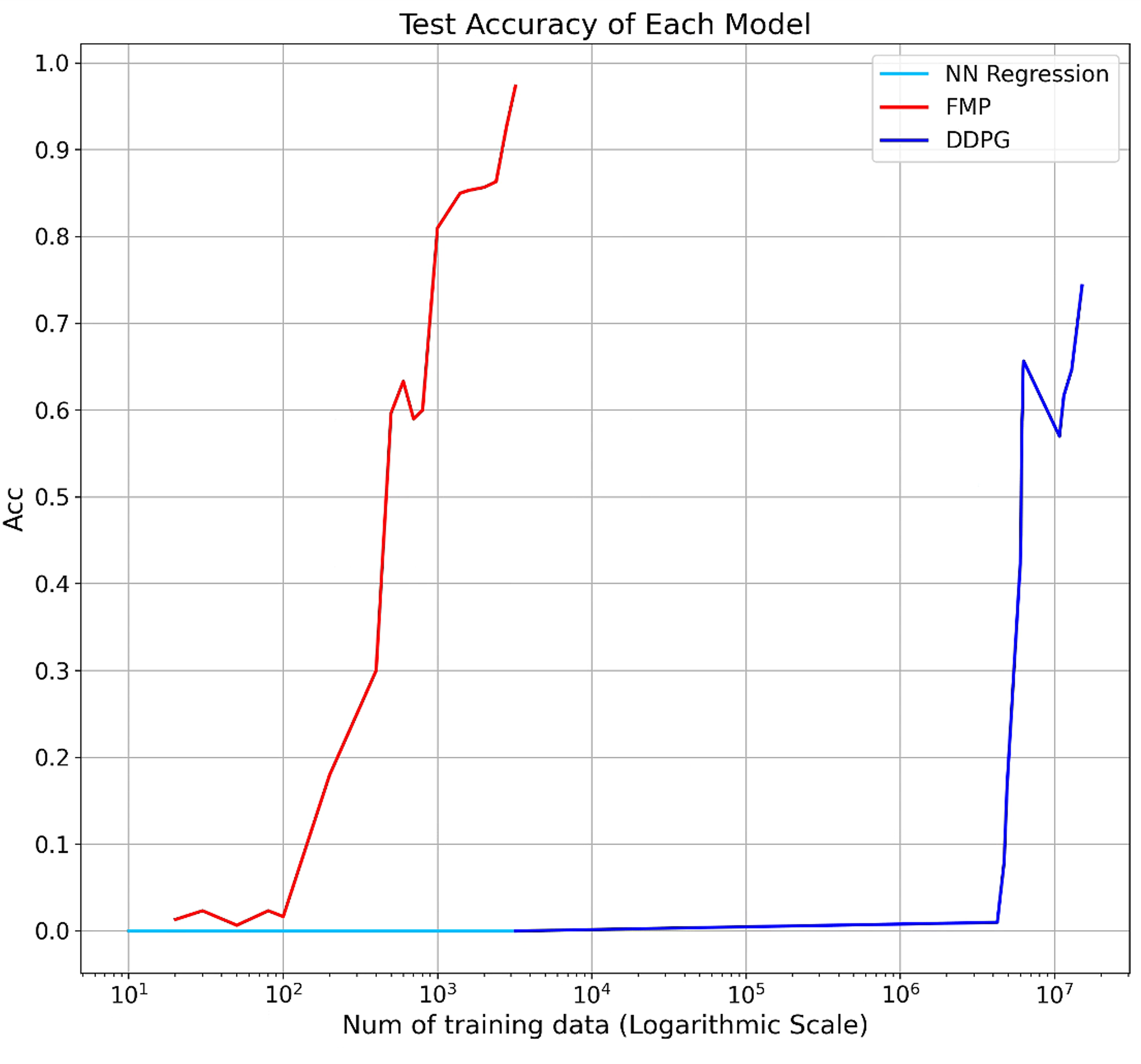}
    \caption{For whole-body control: Test accuracy after training with different data amounts. Supervised learning does not apply to the configuration prediction tasks and cannot give joint angle predictions within the acceptable error $\Delta d$ (Light Blue). We further implemented DDPG to achieve predictions for comparison (Dark Blue)$^{\ref{fn:2},\ref{fn:3}}$.}
    \label{FIG:FMP}
    \vspace{-3mm}
\end{figure}

Figure \ref{FIG:6} and Figure \ref{FIG:FMP} displays the test set results obtained from training different models with varying amounts of data, with the CMP achieving the highest test accuracy at 95.67\% and the FMP achieving 98\%.
As we defined the accuracy metric earlier, supervised learning methods, such as NN Regression, do not perform well in terms of accuracy for the chassis motion prediction tasks.
In the comparative experiment with the CMP, NN Regression needs larger data support in training its parameters in a more optimal direction. The existing volume of data falls short of meeting the parameter refinement requirements for NN Regression, resulting in a test accuracy of approximately 50\%. In contrast, the CMP demonstrates an impressive accuracy exceeding 90\% with only 1000 sets of training data.

Meanwhile, in experiments related to FMP (Figure \ref{FIG:FMP}), we found that NN Regression is not suitable for such configuration prediction tasks. We prepared over 5,000 sets of input values and corresponding labels, adhering to kinematic constraints for supervised training, hoping that NN Regression would learn the underlying implicit whole-body control mapping represented by the dataset.
Rigorous experimental results showed that while NN Regression could learn simple position mapping relationships, it was unable to predict the nine valid parameter configurations required for whole-body control in a single instance(in Eq. \ref{FMP9output}), particularly within the required error(in Eq. \ref{FMP_error}). This further demonstrates that supervised learning is not well-suited to solving sampling problems, as there is no fixed functional mapping for valid solutions. \footnote{In Figure \ref{FIG:FMP}, the light blue represents the NN regression method. We tested six different network architectures, three optimizers, and three kinematic representation relationships, each under 20 different data quantities.\label{fn:2}}

We further attempted to use the Deep Deterministic Policy Gradient(DDPG) method \cite{DDPG} from DRL as a replacement for NN regression to address this prediction problem. The substantial data requirement is justified by our goal for the agent to learn the underlying real-world constraints. By training an Actor-Critic (AC) architecture, we aim to maximize the expected reward in the end. This approach is fundamental to how deep reinforcement learning addresses the challenges associated with sampling in continuous solution spaces. In scenarios characterized by non-fixed mapping relationships, the strategy involves converging the policy space to seek the optimal solution. We constructed an agent comprising two actors and two critics, totaling four networks, and utilized up to 15 million data samples, ultimately achieving an accuracy of 74.33\%. 
\footnote{In Figure \ref{FIG:FMP}, the dark blue indicates DDPG, which is an effective offline deep reinforcement learning method designed for continuous control problems. we employed DDPG for predictions after 3,200 data samples, to the maximum data amount of initial data (204800) and resampled replay buffer update data (max 29000 epoch * 512).\label{fn:3}}
In contrast, FMP once again demonstrated remarkable prediction efficiency, being able to output the entire system's parameter configuration in a single forward propagation while satisfying physical constraints.

Another conclusion is evident from Figure \ref{FIG:6} and Figure \ref{FIG:FMP}: the predictors defined by RobKiNet demonstrate exceptional data efficiency. Compared to NN regression or deep reinforcement learning methods, CMP requires only 1/71 of the data, and FMP only 1/15052 of the data to achieve the same prediction accuracy. This excellent data efficiency can be attributed to the integration of differentiated kinematics algorithms as prior knowledge within the framework.

\begin{figure}[t]
    \centering
    \includegraphics[scale=0.22]{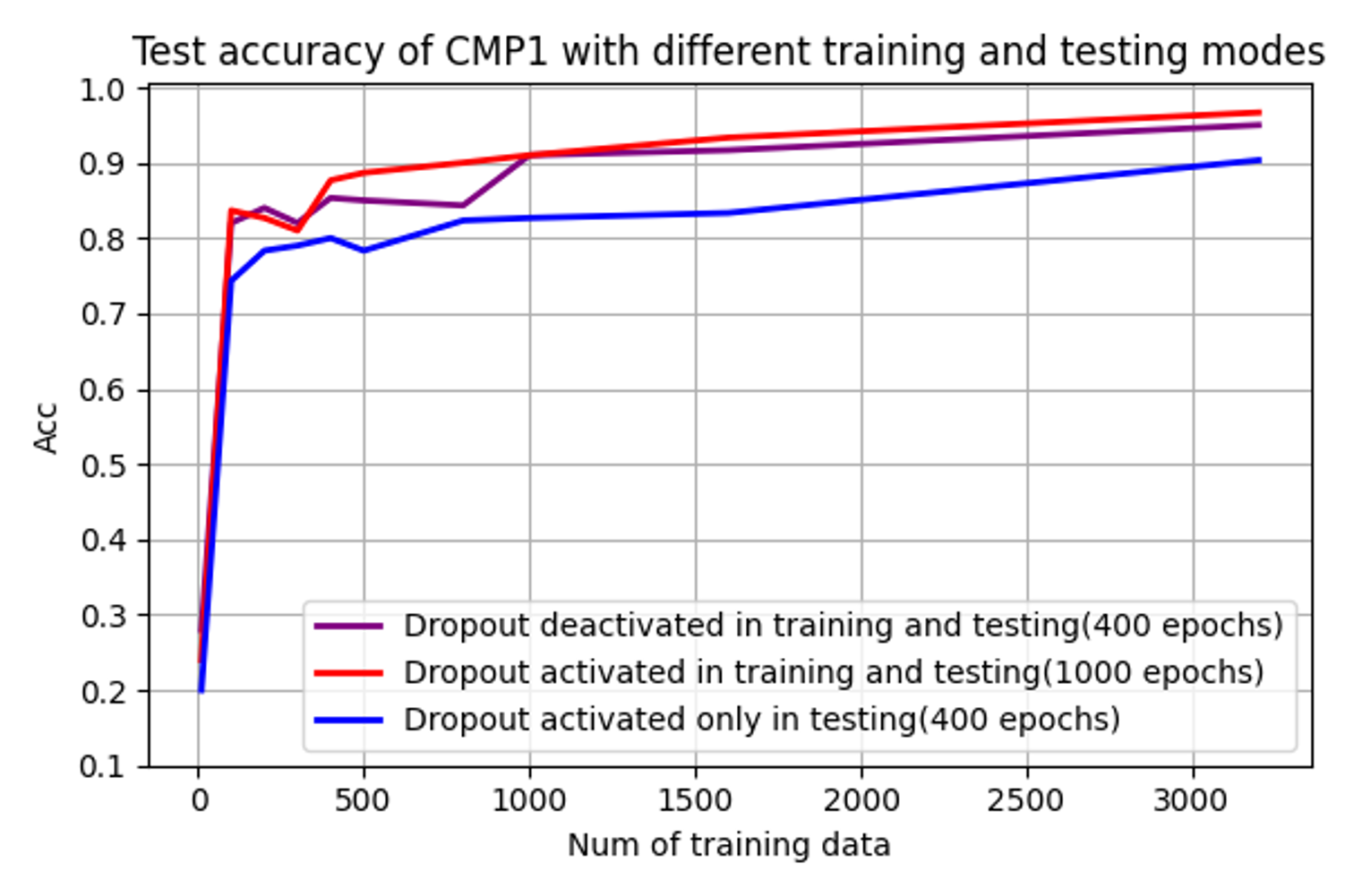}
    \caption{Comparative experiments on CMP1 with the introduction of stochasticity methods.}
    \label{FIG:9}
    \vspace{-3mm}
\end{figure}

\begin{figure}[h]
    \centering
    \includegraphics[scale=0.145]{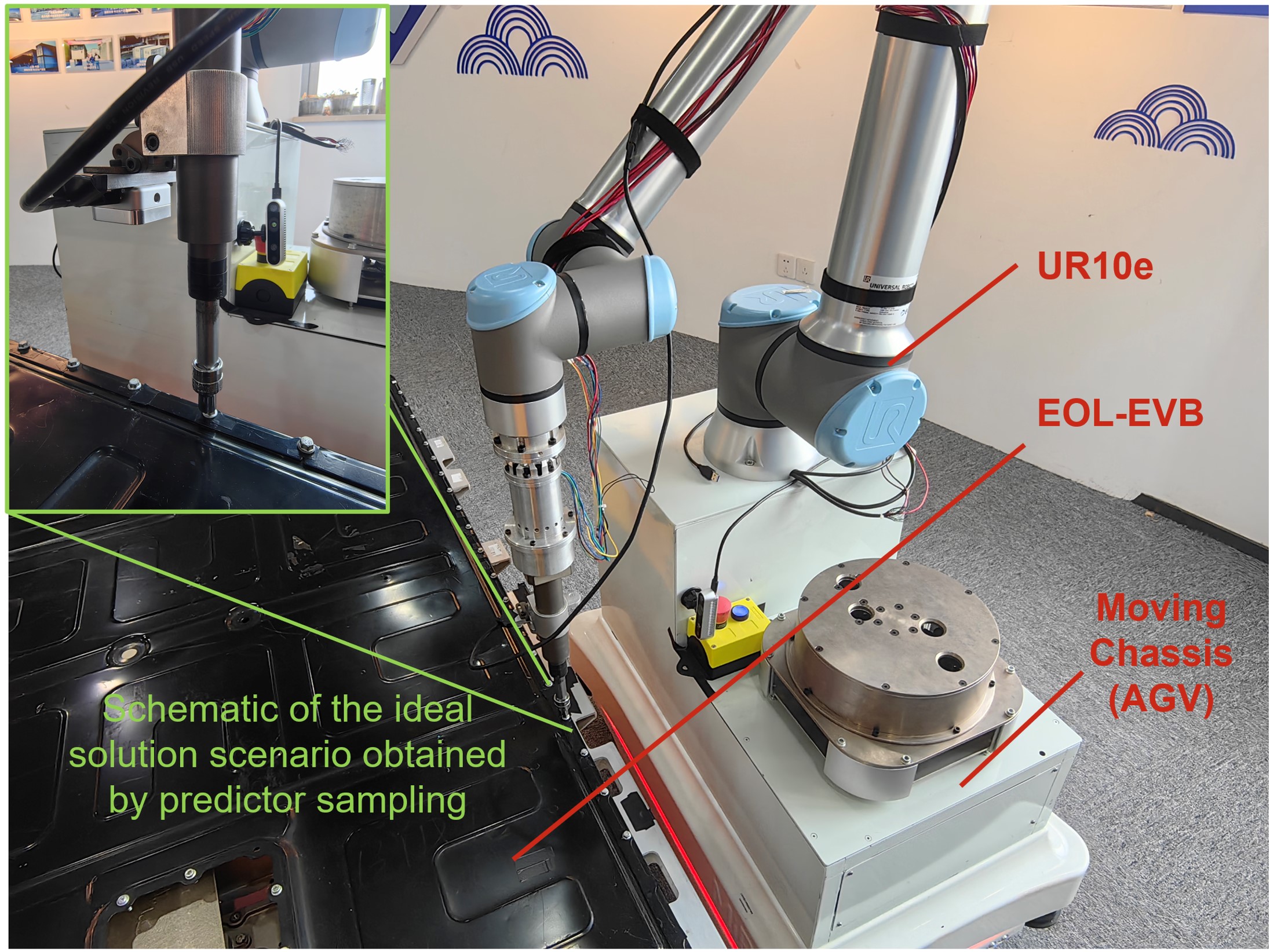}
    \caption{Hardware platform}
    \label{FIG:7}
    \vspace{-3mm}
\end{figure}

\begin{figure*}[h]
    \centering
    \includegraphics[scale=0.12]{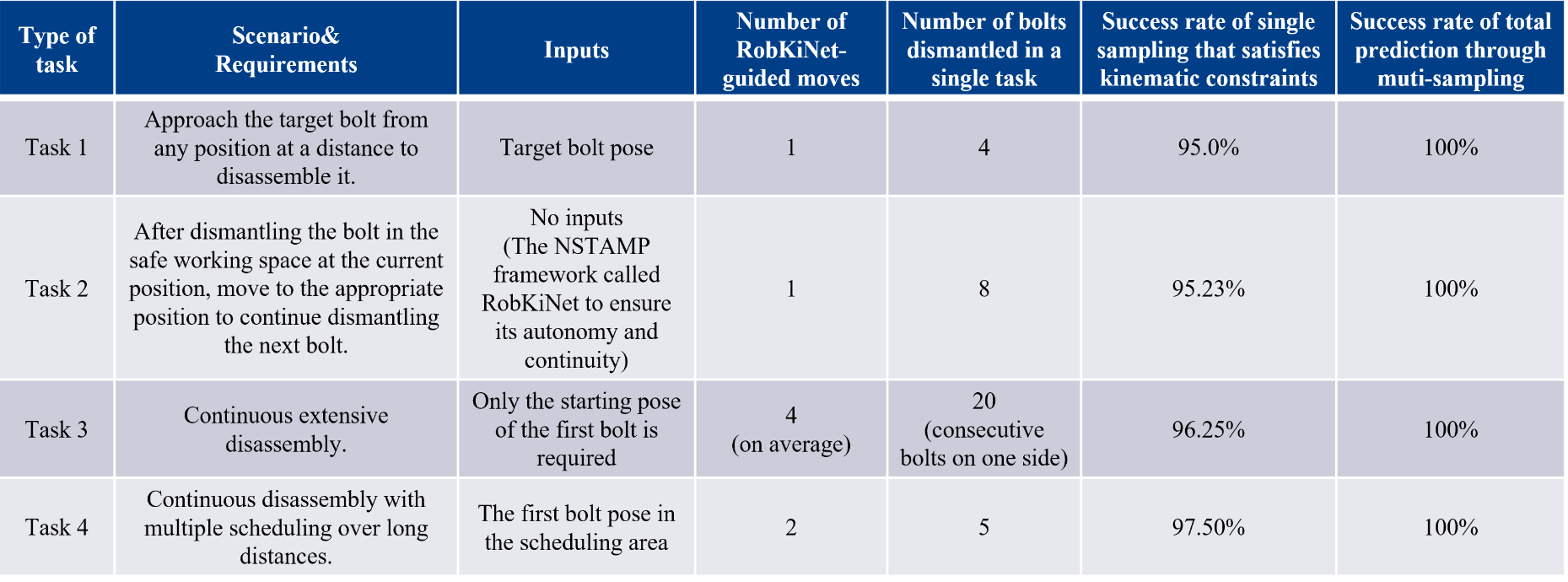}
    \caption{Real-world experiments information and results.}
    \label{FIG:8}
    \vspace{-3mm}
\end{figure*}

It is important to emphasize that the experimental results depicted in Figures 
\ref{FIG:6} represent explorations into the fundamental performance of the predictors, undertaken without the introduction of stochasticity. Compared to RS methods, the CMP and the FMP show amazing advantages in small data sizes and fewer epochs of the training process. However, when undesirable situations below 5\% occur, the fixed model requires multiple network models that have completed training to work together to ensure a 100\% task completion rate. Thus, we introduce the network stochasticity methods described in Section \ref{methods} for a more in-depth comparative experimental exploration by controlling the degree of stochasticity during network training and testing. In this way, we aim to achieve a 100\% success rate by relying solely on the combination of a single pre-trained predictor with the feedback mechanism.
Table \ref{table:3} shows the comparison experiments conducted with CMP1 as the base network. 
Figure \ref{FIG:9} shows the performances of the three methods with different datasets.

Through experiments, it is found that when the stochasticity method is introduced into the training and testing process, the accuracy has more obvious fluctuations in adjacent epochs. The reason for this is that the dropout's culling of hidden neurons is completely randomized, which prevents the network from overly relying on certain neurons to get the desired output. Therefore, when stochasticity is involved in training, the CMP needs to go through more rounds to achieve a similar accuracy as in case 1, and the introduction of stochasticity makes the individual hidden neurons more homogeneous. The experimental results show that after introducing the stochasticity method proposed in this paper, the CMP1 achieves a single prediction accuracy of 96.67\%, but it takes a longer time and training epochs to reach convergence. The accuracy of this method is significantly higher than that of the method in case 3 where stochasticity is introduced only in testing. Therefore, despite the fact that the addition of stochasticity increases the training epoch, we still prefer this approach in our deployments


To further validate the advantages of the RobKiNet, we provided identical disassembly bolt poses for different methods. Except for the networks, we choose basic sampling methods random sampling(RS) \cite{pddlstream} and efficient biased sampling(EBS)\cite{1232271} for clearer comparison with CMP (results in Table \ref{table:4}). We also combine these sampling methods with the traditional Jacobian iteration (Jacobian matrix method)\cite{chen2017tracking} to generate solutions suitable for whole-body control and compare the performance with FMP (results in Table \ref{FMP-Jacobian}). 
We present 300 sets of data and compute the mean sampling times, the sampling time costs, and the time-reduction factor. Compared to random sampling, the CMP achieved a 24.24-fold time reduction, while the FMP achieved a 153-fold time reduction.

\subsection{Real Device Experiments in Disassembly Scenarios}

Figure \ref{FIG:7} is our experimental hardware platform, which consists of an AGV trolley with its mounted 6-axis collaborative robot UR10e working on a to-be-disassembled EOL-EVB.

We deployed the CMP and the FMP for experiments in real-world scenarios. 
The RobKiNet is utilized in the NeuroSymbolic TAMP framework \cite{zhanghengwei2} and acts as the primitive "Approach" in guiding the entire autonomous movement of the AGV chassis. To comprehensively assess the performance of RobKiNet across various usage scenarios, we formulated four tasks. The corresponding task scenarios and requirements are illustrated in Figure \ref{FIG:8}. Through over 80 times of real-machine experiments, we observed that the RobKiNet accurately samples positions in actual disassembly scenarios with a 100\% success rate. 
Demonstration videos are accessible on the paper's website $^{ \ref{fn:supplementary}}$.

\section{Conclusions}

In the framework of robot TAMP, optimizing the sampling efficiency of robot configuration parameters is critical, as it directly impacts the overall efficiency of both task and motion planning while also influencing system robustness. Current approaches, such as neural network regression and reinforcement learning, show promising performance in some areas. However, they rely heavily on extensive data and struggle with inefficiencies in sampling and simulation validation in continuous spaces. Consequently, a key challenge in this research is to enable efficient, end-to-end prediction of robot configuration parameters that strictly adhere to task planning constraints and significantly enhance subsequent motion planning efficiency. Addressing this issue is especially pressing in cases where the robot has redundant degrees of freedom and an infinite number of possible solutions within a continuous solution space.

To address these challenges, we propose an efficient solution by leveraging differentiable programming techniques to deeply incorporate robotic kinematics into the neural network as physical constraints. By embedding forward and inverse kinematic knowledge directly into the training process, this approach effectively constrains the output solution space and enhances optimization gradients. This integration allows for the training of a highly efficient and robust end-to-end robot configuration prediction model, significantly improving the sampling efficiency of robot configuration parameters. The main findings of this research are as follows:

\begin{enumerate}
\item We propose a novel kinematics-informed neural network, RobKiNet, which functions as an interface connecting task planning with motion planning. RobKiNet outputs configuration parameters (e.g., chassis position of the AMMR and robotic arm joint angles) that not only satisfy the global constraints of task planning but also significantly enhance the efficiency of subsequent motion planning. This approach enables the neural network to replace traditional sampling methods, achieving rapid selection of robot configurations while reducing data dependency, and simultaneously meeting the complex constraints at both the task and motion planning levels. Our method enables the network to genuinely learn intricate constraints rather than merely capturing relationships represented by supervised learning datasets.

\item Based on the RobKiNet framework, we developed the CMP and the FMP. When the pose of a work object is given, these predictors can directly output high-reliability predictions for the chassis position and expected robotic arm joint angles, ensuring a viable kinematic solution. Experiments conducted in an EOL-EVB disassembly scenario validated RobKiNet’s efficient robot configuration prediction capabilities, significantly enhancing the efficiency and robustness of AMMR TAMP.

\item Experimental validation shows that the CMP achieves 96.67\% accuracy in initially determining the target chassis position. The FMP provides initial predictions for both the chassis position and corresponding robotic arm joint angles with 98\% accuracy, maintaining an end-effector pose prediction with a position error within 1 mm. Iterative sampling for both predictors consistently ensures 100\% valid chassis positioning. Compared to random sampling, the CMP achieves over an order of magnitude improvement in configuration parameter sampling efficiency, while the FMP achieves over two orders of magnitude. Additionally, RobKiNet demonstrates substantial data efficiency in training; the CMP requires less than 2\% of the training data, and the FMP requires less than 0.01\% to achieve prediction accuracy comparable to other deep learning approaches, such as traditional Jacobian iteration and DDPG.

\item During the implementation of CMP and FMP, a systematic differentiable strategy for forward and inverse kinematics algorithms was developed, along with methods for differentiating non-differentiable operators within computation graphs. Additionally, network stochasticity and custom loss functions were introduced in a sophisticated manner, establishing a comprehensive technique for embedding kinematic constraints into neural networks. This approach not only successfully integrates robotic kinematic knowledge into neural network training but also demonstrates the immense potential of differentiable programming in the field of robotic TAMP. Moreover, it serves as a valuable reference for the construction and optimization of other differentiable computational frameworks.
\end{enumerate}

The RobKiNet system is not only suitable for various types of mobile manipulators but also supports functional and practical verification by offering downloadable CMP and FMP predictors, allowing users to validate its effectiveness across multiple application scenarios. In the future, we plan to incorporate more robotic knowledge, including dynamics, stability, motion control (e.g., collision detection and dynamic property optimization), into the framework, alongside exploring more comprehensive loss definition methods. This development will significantly enhance robotic intelligence, enabling greater adaptability in complex environments. Simultaneously, it will ensure control safety, optimize resource utilization, and improve overall system robustness, ultimately elevating task performance quality.

\section*{Acknowledgment}
The authors express their sincerest thanks to the Ministry of Industry and Information Technology of China for financing this research within the program "2021 High Quality Development Project (TC210H02C)" and support from Neuro-Symbolic Al HOME Satellite Laboratories.



\bibliographystyle{unsrt} 
\bibliography{ref}

\end{document}